\documentclass[10pt,twocolumn,letterpaper]{article}

\usepackage{cvpr}
\usepackage{times}
\usepackage{graphicx}
\usepackage{amsmath,amsthm}
\usepackage{amssymb,empheq}
\usepackage{latexsym}
\usepackage{url}
\usepackage{multirow}

\usepackage[labelfont=bf,font=scriptsize]{subcaption}

\usepackage{algorithm}
\usepackage{algorithmic}

\DeclareGraphicsExtensions{.pdf,.png,.jpg}

\graphicspath{{./}{./Fig/}{./fig/}{./Figs/}{./figs/}}

\newcommand{\BY}{{\mathbf{Y}}}
\newcommand{\BX}{{\mathbf{X}}}
\newcommand{\BD}{{\mathbf{D}}}
\newcommand{\BL}{{\mathbf{L}}}
\newcommand{\Bone}{{\mathbf{1}}}
\newcommand{\BU}{{\mathbf{X}}}

\newcommand{\BB}{{\mathbf{B}}}
\newcommand{\BR}{{\mathbf{X}}}	
\newcommand{\BW}{{\mathbf{W}}}

\newcommand{\BZ}{{\mathbf{Z}}}
\newcommand{\BM}{{\mathbf{M}}}

\newcommand{\BT}{{\mathbf{T}}}
\newcommand{\BI}{{\mathbf{I}}}
\newcommand{\T}{{\!\top}}
\newcommand{\Bx}{{\mathbf{x}}}
\newcommand{\By}{{\mathbf{y}}}

\newcommand{\IDH}{{IMH}\xspace}
\newcommand{\idh}{{Inductive Manifold-Hashing}\xspace}

\renewcommand{\Lambda}{\varLambda}

\newcommand{\st}{{\,\,\mathrm{s.t.\,\,}}}

\newcommand{\diag}{{\mathrm{diag}}}

\newcommand{\trace}{{\mathrm{trace}}}

\newcommand{\w}{{\mathrm{w}}}
\newcommand{\sgn}{{\mathrm{sgn}}}

\ifx\theorem\undefined
\newtheorem{theorem}{Theorem}
\newenvironment{theorem*}{\par\noindent{\bf Theorem\ }}{\hfill\\[2mm]}

\newtheorem{corollary}[theorem]{Corollary}
\newenvironment{corollary*}{\par\noindent{\bf Corollary\ }}{\hfill\\[2mm]}
\newtheorem{definition}[theorem]{Definition}

\fi

\newcommand{\RR}{\mathbb{R}}
\newcommand{\NN}{\mathbb{N}}

\DeclareMathOperator{\Ncal}{\mathcal{N}}
\newcommand{\Bc}{{\mathbf{c}}}

\cvprfinalcopy %
\ifcvprfinal\fi

\usepackage{cite}

\usepackage{eucal}

\begin{document}

\title{Inductive Hashing on Manifolds}

\author{Fumin Shen$^{\ddagger\diamond}$\thanks{
 F. Shen's contribution was made
when he was visiting University of Adelaide.
}  ~ Chunhua
Shen$^{\diamond} $\thanks{
 \it Appearing in IEEE Conf.\ Computer Vision and Pattern Recognition,
 2013. 
 Correspondence should be addressed to C. Shen.
 }  ~ Qinfeng Shi$^{\diamond} $
 ~ Anton van den Hengel$^{\diamond} $
~ Zhenmin Tang$^{\ddagger}$
\\
$^\diamond$ The University of Adelaide, Australia 
~
$^\ddagger$ Nanjing University of Science and Technology, China
}

\maketitle
\thispagestyle{empty}

\begin{abstract}

    Learning based hashing methods have attracted considerable
    attention due to their ability to greatly increase the scale at
    which existing algorithms may operate.
    Most of these methods are designed 
    to generate binary codes 
    that preserve the Euclidean distance in the original space. 
    Manifold learning techniques, in contrast,
    are better able to model the intrinsic structure 
    embedded in the original high-dimensional data.  
    The complexity of these models, and the problems with
    out-of-sample data, have previously rendered them unsuitable for
    application to large-scale embedding, however.

    In this work, we consider how to learn compact binary embeddings
    on their  intrinsic manifolds.
    In order to address the above-mentioned difficulties, 
    we describe an efficient, inductive solution to the 
    out-of-sample data problem, and a process by which 
    non-parametric manifold learning may be used as the basis of a hashing method.
    Our proposed approach thus allows the development of a range of new hashing techniques
    exploiting the flexibility of the wide variety of manifold learning approaches available.
    We particularly show that hashing on the basis of t-SNE 
    \cite{tSNE2008}
    outperforms state-of-the-art hashing methods on
    large-scale benchmark datasets, and %
    is very effective for image classification with very short code
    lengths. 

\end{abstract}

\vspace{-.58cm}
\section{Introduction}

    One of many challenges emerging from the current explosion in the 
    volume of image-based media available is how to
    index and organize the  data accurately, but also efficiently.
    Various hashing techniques have attracted considerable attention 
    in computer vision, information retrieval
    and machine learning \cite{LSH99,SH08,PCA-ITQ,SSH2012,ICML13SHEN}, 
    and seem to offer great promise towards this goal.
    Hashing methods aim to encode documents or images
    as a set of short binary codes, while maintaining aspects of the structure 
    of the original data.
    The advantage of these compact binary representations is that 
    pairwise comparisons may be carried out extremely efficiently.  
    This means that many algorithms which are based on such pairwise 
    comparisons can be made more efficient, and applied to much larger datasets.

Locality sensitive hashing (LSH) \cite{LSH99}  is one of the most well-known
{\it data-independent} hashing methods, and generates  hash codes based on
random projections. 
With the success of LSH, random hash functions have been extended to
 several  similarity measures, including $p$-norm distances
 \cite{LSH_p}, the Mahalanobis metric\cite{kulis2009fast}, and kernel
 similarity \cite{KLSH2009,raginsky2009locality}. However, the methods belonging to the LSH
 family normally require relatively long hash codes and several hash
 tables to achieve both high precision and recall.  This leads to a
 larger storage cost than would otherwise be necessary, and thus limits 
 the sale at which the algorithm may be applied.

{\it Data-dependent} or learning-based hashing methods have 
been developed with the goal of learning more compact hash codes.
Directly learning binary embeddings typically results in an optimization 
problem which is very difficult to solve, however.
Relaxation is often used to simplify the optimization (\eg,
\cite{LDAhash2012,SSH2012}. 
As in LSH, the methods aim to identify a set of hyperplanes, 
but now these hyperplanes are learned, rather than randomly selected.
For example, PCAH \cite{SSH2012}, SSH \cite{SSH2012}, and ITQ
\cite{PCA-ITQ} generate linear hash functions through simple PCA
projections, while LDAhash \cite{LDAhash2012} is based on LDA.
Extending this idea, there are also methods which learn hash functions
in a kernel space, such as reconstructive embeddings
(BRE) \cite{BRE2009}, random maximum margin hashing (RMMH)
\cite{RMMH2011} and kernel-based supervised hashing (KSH)
\cite{KSH2012}.  In a  departure from such methods,  however, spectral
hashing (SH) \cite{SH08}, one of the most popular learning-based
methods, generates hash codes by 
    solving the relaxed mathematical program that is similar to the
    one in Laplacian eigenmaps \cite{LE2001}. 

\begin{figure*}[t]
\centering
\begin{align*}
\begin{subfigure}{0.07\textwidth}
\centering
\includegraphics[height = 1.3cm]{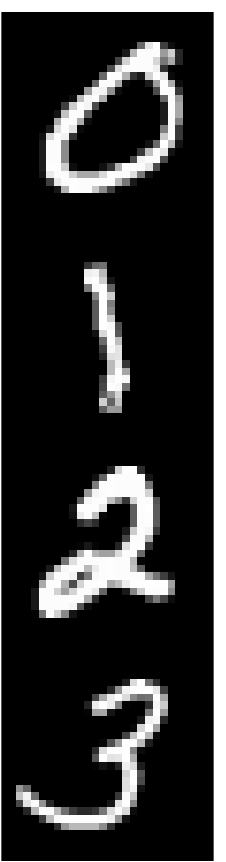}
\caption{Queries}
\end{subfigure} %
\begin{subfigure}{0.2\textwidth}
\includegraphics[height = 1.3cm]{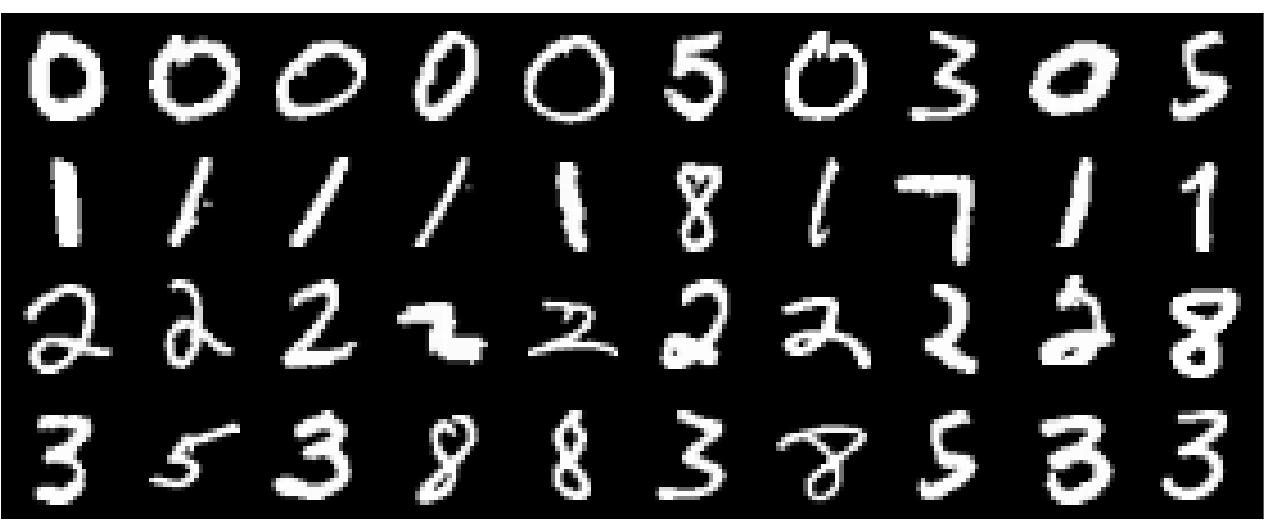}
\caption{$ \ell_2$ dist.\ on 784D}
\end{subfigure} %
\begin{subfigure}{0.2\textwidth}
\includegraphics[height = 1.3cm]{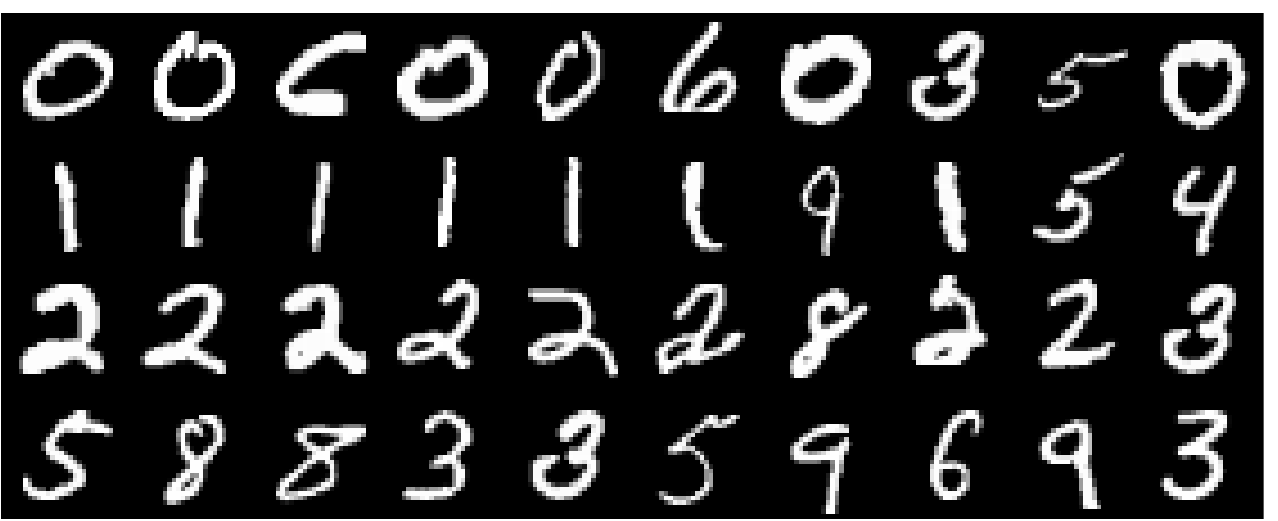}
\caption{LSH with 128-bits}
\end{subfigure}
\begin{subfigure}{0.2\textwidth}
\includegraphics[height = 1.3cm]{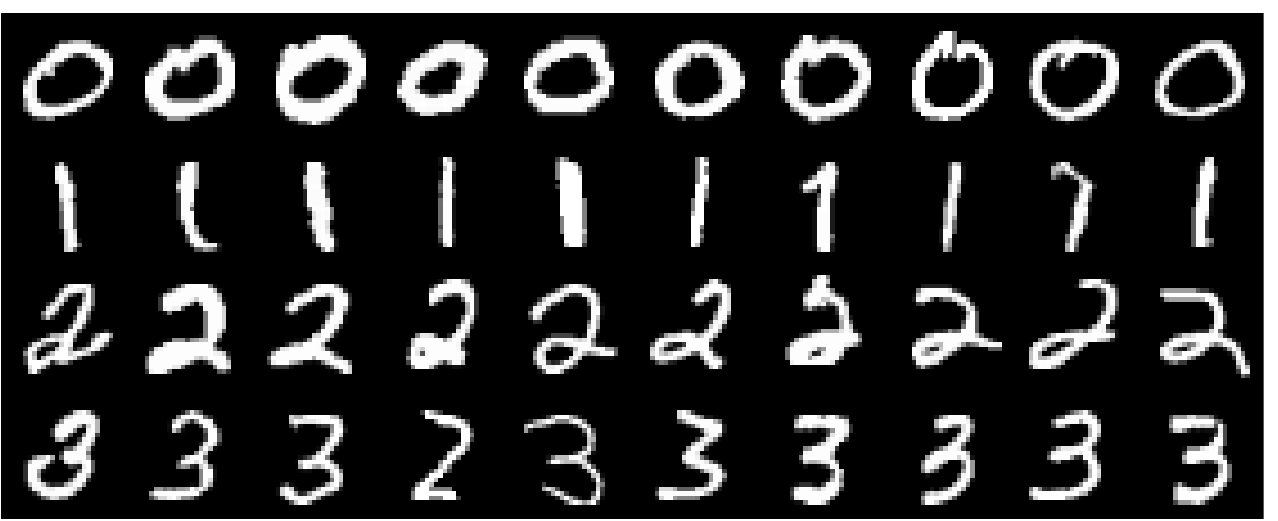}
\caption{$ \ell_2$ dist.\ on embeded 48D}
\end{subfigure} %
\begin{subfigure}{0.2\textwidth}
\includegraphics[height = 1.3cm]{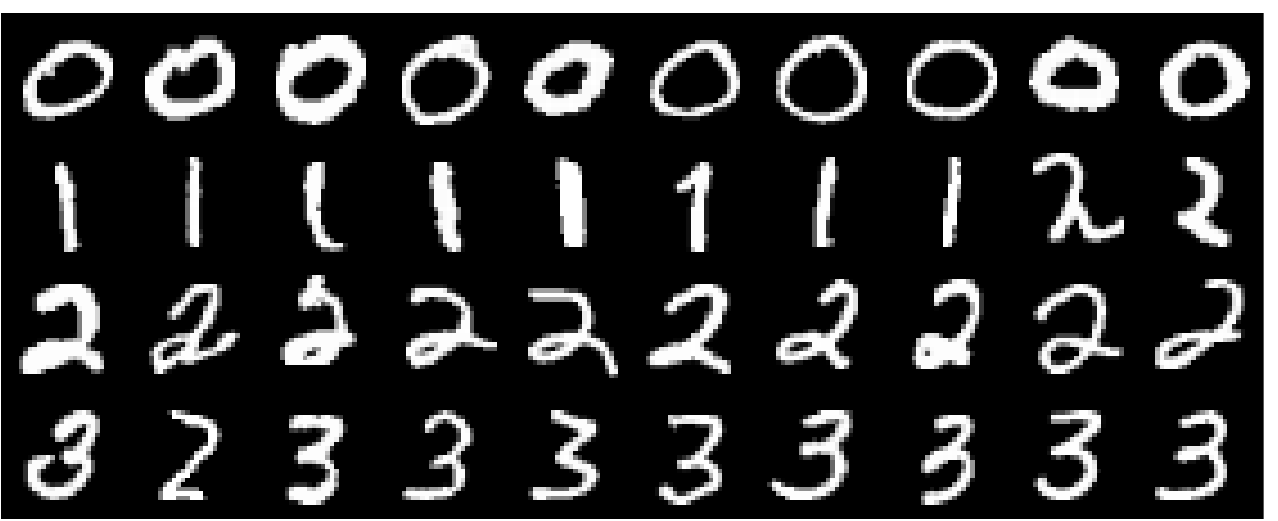}
\caption{Hamming dist.\ with 48-bits}
\end{subfigure}
\end{align*}
\vspace{-0.4cm}
\caption{Top 10 retrieved digits for 4 queries (a) on a subset of MNIST with  300 samples. Search is conducted in the original feature space (b, c) and embedding space by t-SNE\cite{tSNE2008} 
(d, e) using Euclidean distance (b, d) and hamming distance (c, e). }
\label{query_digit}
\end{figure*}

Embedding the original data into a low dimensional space while
simultaneously preserving the inherent neighborhood structure is
critical for learning compact, effective hash codes.
In general, nonlinear manifold learning  methods are more
powerful than  linear dimensionality reduction techniques, as they are
able to more effectively preserve the \textit{local} structure of the
input data without assuming \textit{global} linearity
\cite{talwalkar2008large}. 
The geodesic distance on a manifold has been shown to outperform the Euclidean distance in the high-dimensional  space for image retrieval \cite{He_manifold_2004}, for example.
Figure~\ref{query_digit} demonstrates that searching using 
either the Euclidean or Hamming distance after nonlinear embedding results in more semantically accurate 
neighbors than the same search in the original feature space, and thus
that low-dimensional embedding may actually improve retrieval or
classification performance.
However,
the only widely used nonlinear embedding method for hashing is Laplacian eigenmaps (LE)
(\eg, in \cite{SH08,STH2010,AGH2011}).
Other effective manifold learning approaches (\eg, LLE \cite{LLE2000},
elastic embedding \cite{EE2010} or t-SNE \cite{tSNE2008}) have rarely been explored for hashing.

One problem hindering the use of manifold learning for hashing is that
these methods do not directly scale to large datasets. For example, to
construct the neighborhood graph (or pairwise similarity matrix) in
these algorithms for $n$ data points is $\mathit{O}(n^2)$ in time,
which is intractable for large datasets. 
The second problem is that they are typically non-parametric and thus
cannot efficiently solve the critical out-of-sample extension problem.
This fundamentally limits their application to hashing, as generating
codes for new samples is an essential part of the problem.   One of
the widely used solutions for the methods involving spectral
decomposition (\eg, LLE, LE and ISOMap \cite{ISOMAP2000}) is the
Nystr\"om extension \cite{Nystrom2004}, which solves the problem by
learning  eigenfunctions of a kernel matrix. As mentioned in
\cite{SH08}, however, this is impractical for large-scale hashing
since the Nystr\"om extension is as expensive as doing exhaustive
nearest neighbor search ($\mathit{O}(n)$). A more significant problem,
however,  is the fact that the Nystr\"om extension cannot be directly
applied to {\it non-spectral} manifold learning methods such as t-SNE.

In order to address the out-of-sample extension problem, we propose a
new non-parametric regression approach which is both efficient and
effective.  This method allows rapid assignment of new codes to
previously unseen data in a manner which preserves the underlying
structure of the manifold.  Having solved the out-of-sample extension
problem, we develop a method by which a learned manifold may be used
as the basis for a binary encoding.  This method is designed so as to
generate encodings which reflect the geodesic distances along such
manifolds.  On this basis we develop a range of new embedding
approaches based on a variety of manifold learning methods.  The best
performing of these is based on manifolds identified through t-SNE,
which has been shown to  be effective in discovering semantic
manifolds amongst the set of all images\cite{tSNE2008}.

Given the computational complexity of many manifold learning methods, we 
show that it is possible to learn the manifold on the basis of a small subset of the data $\BB$ (with size $m$), and subsequently to inductively insert the remainder of the data, and any out-of-sample data, into the embedding in $\mathit{O}(m)$ time per point.
This process leads to an embedding method we label \idh (\IDH) which we show to
outperform state-of-the-art methods on several large scale datasets both quantitatively and qualitatively.

\paragraph{Related work}
\textit{Spectral Hashing}
Weiss et al. \cite{SH08} formulated the spectral hashing (SH) problem as
\begin{align}
\label{EQ:SH}
\min_\BY & \sum_{\Bx_i,\Bx_j \in \BU} \w(\Bx_i, \Bx_j) \|\By_i -\By_j\|^2
\\ 
\st & \BY \in \{-1,1\}^{n \times r}\notag, \, 
  \BY^{\T} \BY = n\BI \notag, \,
  \BY^\T \Bone = 0. \notag 
\end{align}
Here $\By_i \in \{-1, 1\}^r$, the $i$th row in $\BY$, is the hash code we want to learn for $\Bx_i \in \mathbf{R}^d$, which is one of the $n$ data points in the training data set $\BU$. 
$\BW \in \mathbf{R}^{n \times n}$ with $\BW_{ij} = \w(\Bx_i, \Bx_j) = \exp(-\|\Bx_i - \Bx_j\|^2/\sigma^2)$ is the graph affinity matrix, where $\sigma$ is the bandwidth parameter.  
 $\BI$ is  the identity matrix. 
The last two constraints force the learned hash bits to be uncorrelated and balanced, respectively.
By removing the first constraint (\ie, \textit{spectral relaxation}\cite{SH08}),  $\BY$ can be easily obtained by 
spectral decomposition on the Laplcaian matrix $\BL =
 \BD - \BW$, where $\BD = \diag(\BW \Bone)$ and $\Bone$ is the vector with all ones. 
 However, constructing $\BW$ is $\mathit{O}(dn^2)$ (in time) and calculating the Nystr\"om extension for a new point is $\mathit{O}(rn)$, which are both intractable for large datasets. 
It is assumed in SH \cite{SH08}, therefore, that the data are sampled
from a uniform distribution, which leads to a simple analytical
eigenfunction solution of 1-D Laplacians. However, this strong
assumption is not true in practice and the manifold structure of the
original data are thus destroyed \cite{AGH2011}.

\textit{Anchor Graph Hashing}
To efficiently solve problem \eqref{EQ:SH}, anchor graph hashing (AGH)  \cite{AGH2011} approximated the affinity matrix $\BW$ by the low-rank matrix $\hat{\BW} = \BZ\Lambda^{-1}\BZ$, where $\BZ \in \mathbf{R}^{n \times m}$ is the normalized affinity matrix (with $k$ non-zeros in each row) between the training samples and $m$ \textit{anchors}  
(generated by K-means), and $\Lambda^{-1}$ normalizes $\hat{\BW}$ to be doubly stochastic. 
Then  the desired hash functions may be efficiently identified by binarizing the Nystr\"om eigenfunctions \cite{Nystrom2004} with the approximated affinity matrix $\hat{\BW}$.  AGH 
is thus efficient, in that it has linear training time and constant search time, but 
as is the case for SH \cite{SH08}, the generalized eigenfunction is derived only  for the Laplacian eigenmaps embedding.

\textit{Self-Taught Hashing}
Self-taught hashing (STH) \cite{STH2010} addressed the out-of-sample problem by a novel way:  hash functions are obtained by
training an SVM classifier for each bit using the pre-learned binary codes as class labels. The binary codes were learned by directly solving \eqref{EQ:SH} with a cosine similarity function.
This process has  prohibitive computational and memory costs, however,
and training the SVM can be very time consuming for dense data.

\section{The proposed method}
\subsection{Inductive learning for hashing}
Assuming that we have the  manifold-based embedding $\BY := \{\By_1,
\By_2, $ $ \cdots, $ $ \By_n\}$ for the entire training data $\BX:=
\{\Bx_1, $ $ \Bx_2, $ $\cdots, $ $\Bx_n\}$.
Given a new data point $\Bx_q$, we aim to 
generate an
embedding $\By_q$ which preserves the local neighborhood relationships 
among its neighbors $\mathcal{N}_k(\Bx_q)$ in $\BX$.
We choose to minimize the following simple objective:
\begin{equation}
\label{EQ:newdata}
\mathcal{C}(\By_q) = \sum_{i = 1}^n \w(\Bx_q, \Bx_i) \|\By_q -\By_i\|^2.
\end{equation}
Here we define
\begin{align*}
\w(\Bx_q, \Bx_i) = 
\left\{
\begin{array}{cl}
\exp (-\|\Bx_q - \Bx_i\|^2/\sigma^2), & \text{if } \Bx_i \in \mathcal{N}_k(\Bx_q),\\
0 & \text{otherwise}.
\end{array}\right.
\end{align*}
Minimizing \eqref{EQ:newdata} naturally 
uncovers an embedding for the new point on the basis of its nearest neighbors on the
low-dimensional manifold initially learned on the base set.
That is, in the low-dimensional space, the new embedded location for the point should be close to those of the points close to it in the original space.

Differentiating $\mathcal{C}(\By_q)$ with respect to $\By_q$, we have
\begin{equation}
\frac{\partial \mathcal{C}(\By_q)}{\By_q}\bigg|_{\By_q = \By_q^{\star}} = \,2\sum_{i = 1}^n \w(\Bx_q, \Bx_i) (\By_q^{\star} -\By_i) = 0,
\end{equation}
which leads to the optimal solution
\begin{equation}
\label{EQ:Induction}
\By_q^{\star} = \frac{\sum_{i = 1}^n \w(\Bx_q, \Bx_i) \By_i}{\sum_{i = 1}^n \w(\Bx_q, \Bx_i)}.
\end{equation}
Equation \eqref{EQ:Induction} provides a simple inductive formulation for the embedding: produce the embedding for a new data point by a (sparse) linear combination of the base embeddings. 

The proposed approach here is inspired by 
        Delalleau et al.\  \cite{Olivier2005}, where they have focused
        on non-parametric graph-based learning in  semi-supervised
        classification. Our aim here is completely different: We try
        to scale up the manifold learning process for hashing in an
        unsupervised manner.
 
 The resulting solution \eqref{EQ:Induction} is consistent with the basic 
smoothness
 assumption in manifold learning, that close-by data points lie on or close to a locally linear manifold \cite{LLE2000,ISOMAP2000,LE2001}.
 This local-linearity  assumption has also been widely used in semi-supervised learning \cite{Olivier2005,LLC2009},  image coding \cite{wang2010locality}, and similar. 
In this paper, we propose to apply this assumption to hash function learning.

However, as aforementioned,  \eqref{EQ:Induction} does not scale well
for both computing $\BY$ ($\mathit{O}(n^2)$ \eg, for LE) and
out-of-sample extension ($\mathit{O}(n)$), which is intractable for
large scale tasks. 
Next, we  show that the following prototype algorithm is able to approximate $\By_q$ using only a small base set well. 
 This prototype algorithm is based on entropy numbers defined below.

\begin{definition} [Entropy numbers \cite{HerWil02c}] Given any $\BY \subseteq \RR^r$ and $p \in \NN$, the $m$-th entropy number $\epsilon_m(\BY)$ of $Y$ is defined as
\[\epsilon_{m}(Y) := \inf\{\epsilon>0| \Ncal (\epsilon, Y, \|\cdot - \cdot \|) \leq m \},\]
where $\Ncal$ is the covering number. Then $\epsilon_{m}(Y)$ is the smallest radius that $Y$ can be covered by less or equal to $m$ balls.
\end{definition}

\subsubsection{The prototype algorithm}
Inspired by Theorem 27 of \cite{HerWil02c}, we construct a prototype algorithm below. 
 We use $m$ clusters to cover $\BY$. Let $\alpha_i =\frac{ \w(\Bx_q, \Bx_i) }{\sum_{j =1}^n \w(\Bx_q, \Bx_j)} $ and $C_j = \sum_{i\in I_j} \alpha_i$.  For each cluster index set $I_j$, we randomly draw $\ell_j = \lfloor mC_j+1 \rfloor$ many indices from $I_j$ proportional to their weight $\alpha_i$. 
 That is, for $\mu \in \{1,\cdots, \ell_j\}$,  the $\mu$-th randomly drawn index  $u_{j,\mu}$ 
$ \Pr ( u_{j,\mu} = i) = \frac{\alpha_i}{C_j}, \forall j \in \{1,\cdots,m\}.$
 We then construct $\hat{\By}_q$ as
 \begin{align}
 \label{eq:infer-sample}
 \hat{\By}_q = \sum_{j=1}^m \frac{C_j}{\ell_j} \sum_{\mu=1}^{\ell_j}\By_{u_{j,\mu}}.
 \end{align}

   \begin{theorem} 
   \label{thm:appox}
   For any even number $n' \leq n$. If Prototype Algorithm uses $n'$ many non-zero $\By \in \BY$ to express $\hat{\By}_q$, then
  \begin{align}
  \Pr[\|\hat{\By}_q - {\By}_q\| \ge t ] < \frac{2(\epsilon_{\frac{n'}{2}}(\BY))^2}{n't^2}.
  \end{align}
 \end{theorem}

\begin{corollary} For an even number $n'$, any $\epsilon > \epsilon_{\frac{n'}{2}}(\BY)$, any $\delta \in (0,1)$ and any $t >0$, if $n' \ge  \frac{2\epsilon^2}{\delta t^2}$, then with probability at least $1-\delta$, 
\[\|\hat{\By}_q - {\By}_q\| < t.\]
\end{corollary} 
 Refer to the supplementary material for the  proofs of the  theorem and corollary.
  The quality of the approximation depends on $\epsilon_{\frac{n'}{2}}(\BY)$ and $n'$. If data
  exhibit strong clustering patterns, \ie, data within each cluster are very close to cluster center, we will
  have small $\epsilon_{\frac{n'}{2}}(\BY)$, hence better approximation. Likewise, the bigger $n'$
  is, the better approximation is.

\subsubsection{Approximation of the prototype algorithm}
 The clusters can be obtained via clustering algorithm such as K-means. 
Since the $n$ could be potentially massive, it is impractical to compute  $\alpha_i$ within all clusters. %
Let $\alpha_i(\Bx_q) = \frac{ \w(\Bx_q, \Bx_i) }{\sum_{j =1}^n \w(\Bx_q, \Bx_j)}$. Ideally, for each cluster, we want to select the $\By_i$ that has high overall weight $O_i = \sum_{\Bx_q \in X} \alpha_i(\Bx_q)$. For large scale $\BU$,  we only have limited information available such as cluster centers $\{\Bc_j, j = 1, \cdots, m\}$ and $\w(\Bc_j, \Bx), \Bx \in \BU $. Fortunately, the clustering result gives useful information about $O_i$. 
The cluster centers have the largest overall weight w.r.t the points from their own cluster, \ie $\sum_{{i} \in I_j} { \w({\Bc_j}, \Bx_i) }$. 
This suggests we should select all cluster centers to express $\hat{\By}_q$.

Following many methods in the area (\eg, \cite{SH08,AGH2011}), we obtain our general inductive hash function by  binarizing the low-dimensional embedding 
\begin{equation}
\label{EQ:hash_fun}
h(\Bx) = \sgn\left( \frac{\sum_{j = 1}^m \w(\Bx,\Bc_j) \By_j}{\sum_{j = 1}^m \w(\Bx,\Bc_j)}\right),
\end{equation}
where $\sgn(\cdot)$ is the sign function and $\BY_{\BB}:= $ $\{\By_1,
$ $\By_2, $ $\cdots, $ $ \By_m\}$ is the embedding for 
the base set $\BB $ $ := $ $ \{\Bc_1, $ $ \Bc_2, $ $ \cdots, \Bc_m\}$,
which is the cluster centers obtained by K-means. 
Here we assume that the embeddings $ \By_i $ are centered on the
origin. 
We term our hashing
method \textit{\idh} (\IDH).
The inductive hash function provides a natural means for
generalization to new data, which has a constant $\mathit{O}(dm + rk)$
time.
With this, the embedding for the training data becomes %
\begin{equation}
\BY = \Bar{\BW}_{\BU\BB} \BY_{\BB},
\label{EQ:SubInduction}
\end{equation}
where $\bar{\BW}_{\BU\BB}$ is defined such that $\bar{\BW}_{ij} $ $ =
$ $\frac{\w(\Bx_i, \Bc_j)}{\sum_{i = 1}^m \w(\Bx_i,\Bc_j)}, $ $ \text{for }
\Bx_i $ $\in \BU, $ $ \Bc_j \in \BB$.

Although the objective function \eqref{EQ:newdata} is formally related to LE, it is general in preserving local similarity.
The embeddings $\BY_{\BB}$ can be learned by any appropriate
 manifold learning method which preserves the similarity of interest in the low dimensional space.
We empirically evaluate several other embedding methods in
Section~\ref{Embedding_selection}. Actually, as we show, some manifold learning methods
(\eg,
t-SNE described in Section \ref{SEC:tSNE}) can be better choices for learning binary codes, although LE has been widely used.
We will discuss two methods for learning $\BY_{\BB}$ in the sequel.

We summarize the \idh framework in Algorithm 1. Note that the  computational cost
is dominated by K-means in the first step, which is $\mathit{O(dmnl)}$ in time (with $l$ the number of iterations). Considering that $m$ (normally a few hundreds) is much less than $n$, 
and is a function of manifold complexity rather than the volume of data,
the total training time is linear in the size of training set. 
If the embedding method is LE, for example, then using \IDH to compute $\BY_{\BB}$ requires constructing 
 the small affinity matrix $\BW_{\BB}$ and solving $r$ eigenvectors of the $m \times m$ Laplacian matrix $\BL_{\BB}$
which is $\mathit{O}(dm^2 + rm)$.
Note that in step 3, to compute $\Bar{\BW}_{\BR\BB}$, one needs to compute the distance matrix between $\BB$ and $\BR$, which is a natural output of K-means, or can be computed additionally in $\mathit{O}(dmn)$ time.
The training process on a dataset of 70K items with 784 dimensions can thus be achieved in a few seconds on a standard desktop PC.

\begin{algorithm}[t!]
\caption{\small \idh (\IDH)} 
{
\footnotesize
\textbf{Input: }
 Training data $\BU := \{\Bx_1, \Bx_2, \ldots, \Bx_n\}$, code length $r$, base set size $m$, 
 neighborhood size $k$

\textbf{Output: } Binary codes $\BY := \{\By_1, \By_2, \ldots, \By_n\} \in \mathbb{R}^{n \times r}$

1) Generate the base set $\BB$ by random sampling or clustering (\eg K-means).

2) Embed $\BB$ into the low dimensional space by \eqref{EQ:tSNE}, \eqref{EQ:Obj_main}  or any other 
appropriate manifold leaning method.

3) Obtain the low dimensional embedding $\BY$ for the whole dataset inductively by Equation  \eqref{EQ:SubInduction}.

4) Threshold $\BY$ at zero.
}
\label{Alg1}
\end{algorithm}

\textbf{Connection to the Nystr\"om method} 
As Equation \eqref{EQ:Induction}, the Nystr\"om eigenfunction by
Bengio et al.\ \cite{Nystrom2004} also generalizes to a new point by a linear combination of a set of low dimensional embeddings:
\begin{equation*}
\phi(\Bx) = \sqrt{n}\sum_{j = 1}^{n} \tilde{\mathrm{K}}(\Bx, \Bx_j)\mathbf{V}_{r}^{j}\mathbf{\Sigma}_r^{-1}.
\end{equation*}
For LE,  $\mathbf{V}_r$ and $\mathbf{\Sigma}_r$ correspond to the top
$r$ eigenvectors and eigenvalues of a normalized kernel matrix $\tilde{K}$ with $\tilde{K}_{ij} = \tilde{\mathrm{K}}(\Bx_i, \Bx_j) = \frac{1}{n}\frac{\w(\Bx_i,\Bx_j)}{\sqrt{E_{\Bx}[\w(\Bx_i, \Bx)]E_{\Bx}[\w(\Bx, \Bx_j)]}}$.
In AGH \cite{AGH2011}, the formulated hash function was proved to be the corresponding Nystr\"om eigenfunction with the approximate low-rank affinity matrix.
LELVM \cite{LELVM2007} also formulate out-of-sample mappings for LE in a manner 
similar to \eqref{EQ:Induction} by combining 
latent variable models.
    Both of these methods, and ours, can thus be seen as applications
    of the Nystr\"om method. {\em Note, however, that our method differs in
    that it is not restricted to spectral methods such as LE, and that
    we aim to learn binary hash functions for similarity-based search
    rather than dimensionality reduction.}
    LELVM \cite{LELVM2007} cannot be applied to other embedding
    methods other than LE.

\subsection{Stochastic neighborhood preserving hashing}
\label{SEC:tSNE}
In order to demonstrate our approach we now derive a hashing method based on t-SNE \cite{tSNE2008}, which is a non-spectral embedding method.
t-SNE is a modification of stochastic neighborhood embedding (SNE) \cite{SNE2002} which aims to overcome the tendency of that method to crowd points together in one location.  
 t-SNE provides an effective technique  for visualizing data and dimensionality reduction, which is capable of preserving local structures in the high dimensional data while retaining some global structures \cite{tSNE2008}. These properties make t-SNE a good choice for nearest neighbor search. Moreover, as stated in \cite{venna2010information}, the cost function of t-SNE in fact maximizes the \textit{smoothed recall} \cite{venna2010information} of query points and their neighbors.

The original t-SNE does not scale well, as it has a time complexity which is quadratic in $n$.  More significantly, however, it has a non-parametric form, which means that there is no simple function which may be applied to out-of-sample data in order to calculate their coordinates in the embedded space.
As was proposed in the previous subsection, we 
first apply t-SNE to the base set $\BB$ \cite{tSNE2008},
\begin{equation}
\min_{\BY_{\BB}} = \sum_{\Bx_i \in \BB} \sum_{\Bx_j \in \BB} p_{ij} \log\bigg(\frac{p_{ij}}{q_{ij}}\bigg).
\label{EQ:tSNE}
\end{equation} 
Here $p_{ij}$ is the symmetrized conditional probability in the high dimensional space, and
 $
q_{ij}
$ is the joint probability defined using the t-distribution in the low dimensional embedding space.  
The optimization problem \eqref{EQ:tSNE} is easily solved by a gradient descent procedure.
 After we get embeddings $\BY_{\BB}$ of samples $\Bx_i \in \BB$, the hash codes for the entire dataset can be easily computed %
using~\eqref{EQ:hash_fun}.
It is this method which we label \IDH-tSNE.

\subsection{Hashing with relaxed similarity preservation}
\label{SEC:app_cons}
As in the last subsection, we can compute $\BY_{\BB}$ considering local smoothness only within $\BB$. Based on equation \eqref{EQ:Induction}, in this subsection, we alternatively compute $\BY_{\BB}$ by considering the smoothness both within $\BB$ and between $\BB$ and $\BR$. As in \cite{Olivier2005}, 
the objective can be easily obtained by modifying \eqref{EQ:SH} as:
\begin{align}
\label{EQ:appro_cons}
\mathcal{C}(\BY_{\BB}) & = \sum_{\Bx_i,\Bx_j \in \BB} \w(\Bx_i, \Bx_j) \|\By_i -\By_j\|^2 & (\mathcal{C}_{\BB\BB}) \notag \\ \notag
 & + \lambda \sum_{\Bx_i \in \BB, \Bx_j \in \BR} \w(\Bx_i, \Bx_j) \|\By_i -\By_j\|^2  & (\mathcal{C}_{\BB\BR})\\ 
\end{align}
where $\lambda$ is the trade-off parameter.
$\mathcal{C}_{\BB\BB}$ enforces smoothness of the learned embeddings within  $\BB$ while $\mathcal{C}_{\BB\BR}$ ensures the smoothness between $\BB$ and $\BR$. This formulation is actually a relaxation of \eqref{EQ:SH}, by discarding the  part which minimizes the dissimilarity within $\BR$ (denoted as $\mathcal{C}_{\BR\BR}$). 
$\mathcal{C}_{\BR\BR}$ is ignored since computing the similarity matrix within $\BR$ costs $\mathit{O}(n^2)$ time. The smoothness between points in $\BR$ is implicitly ensured by \eqref{EQ:SubInduction}.

Applying equation  \eqref{EQ:SubInduction} for $\By_j, j \in \BR$ to \eqref{EQ:appro_cons}, we obtain the following problem
\begin{align}
\label{EQ:Objective1}
\min &\, \trace(\BY_{\BB}^\T (\BD_{\BB} - \BW_{\BB}) \BY_{\BB})\\ \notag
 + &\lambda \, \trace(\BY_{\BB}^\T(\BD_{\BB\BR} - \bar{\BW}_{\BR\BB}^\T \BW_{\BR\BB})\BY_{\BB}),
\end{align}
where $\BD_{\BB} = \diag(\BW_{\BB}\Bone)$ and $\BD_{\BB\BR} = \diag(\BW_{\BB\BR}\Bone)$ are both $m \times m$ diagonal matrices.
Taking the constraint in \eqref{EQ:SH}, we obtain 
\begin{align}
\label{EQ:Obj_main}
\min_{\BY_{\BB}} \, & \trace(\BY_{\BB}^\T(\BM + \lambda \BT)\BY_{\BB})\\ \notag
\st & \BY_{\BB}^\T \BY_{\BB} = m\BI
\end{align}
where $\BM = \BD_{\BB} - \BW_{\BB}$, $\BT = \BD_{\BB\BR} - \bar{\BW}_{\BR\BB}^\T \BW_{\BR\BB}$.
The optimal solution $\BY_{\BB}$ of the above problem is easily obtained by identifying the $r$ eigenvectors of $\BM + \lambda\BT$ corresponding to the smallest eigenvalues (excluding the eigenvalue 0 with respect to the trivial eigenvector $\Bone$)\footnote{We set $\lambda$ to 2 in all experiments.}.
We name this method \IDH-LE in the following text.
\subsection{Manifold learning methods for hashing}
\label{Embedding_selection}
\begin{figure}
\centering
\includegraphics[width = 0.45\textwidth]{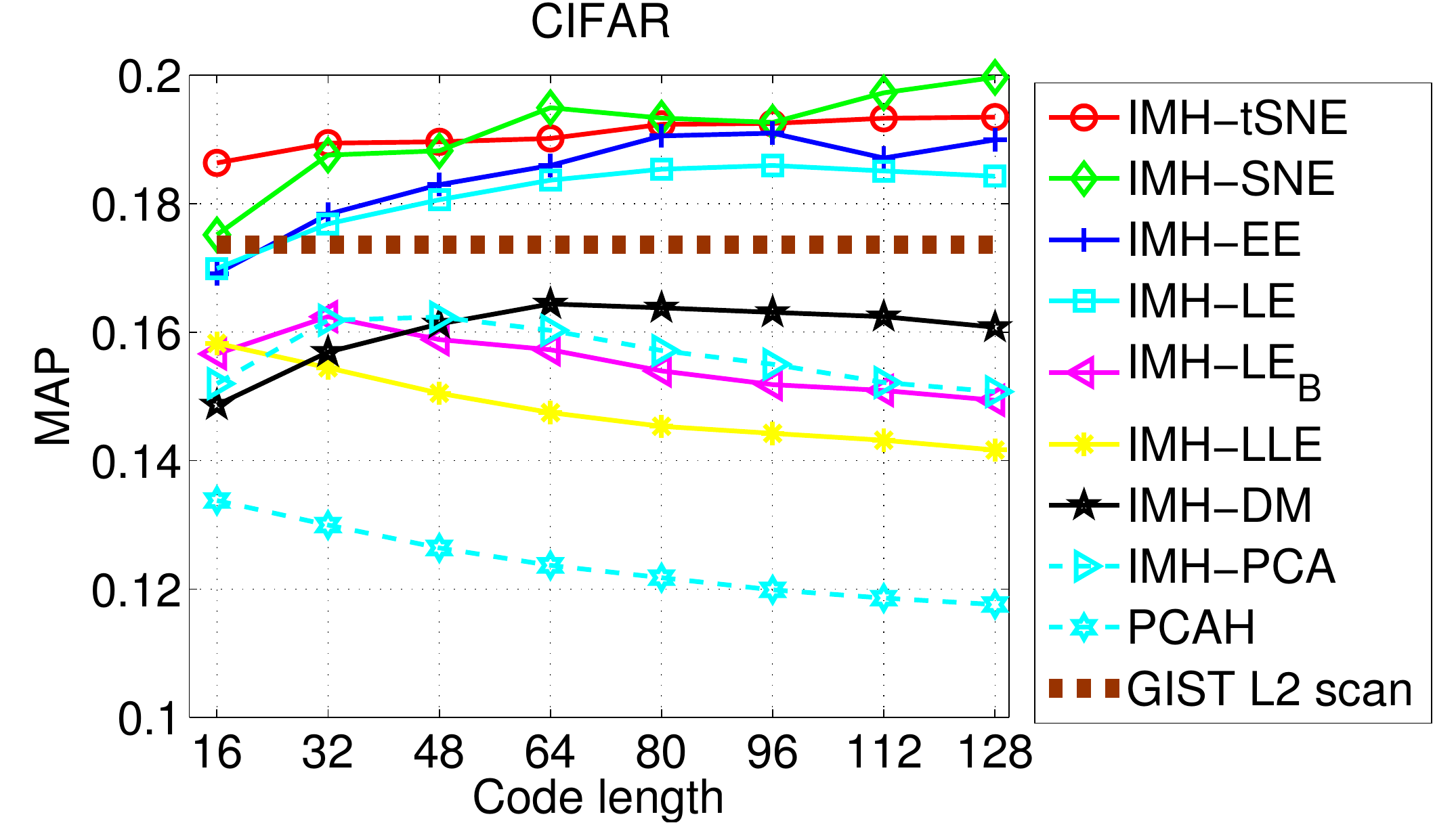}
\caption{Comparison among different manifold learning methods within
our \IDH hashing framework on  CIFAR-10. \IDH with the linear PCA
(\IDH-PCA) and PCAH \cite{SSH2012} are also evaluated for comparison.
For clarity, for\IDH-LE in Section \ref{SEC:app_cons}, we term \IDH with the original LE algorithm on the base set $\BB$ as \IDH-LE$_\BB$.
IMH-DM is IMH with the diffusion maps of \cite{lafon2006diffusion}. 
}
\label{Fig:EmbeddingSelection}
\end{figure}

In this section, we compare different manifold learning methods for hashing within our \IDH framework.
The comparison results are reported in Figure~\ref{Fig:EmbeddingSelection}.
For comparison, we also evaluate the linear PCA within the framework (\IDH-PCA in the figure).
  We can clearly see that \IDH-tSNE, \IDH-SNE and \IDH-EE (with
  Elastic Embedding (EE) \cite{EE2010})  perform slightly better than
  \IDH-LE (Section \ref{SEC:app_cons}). This is mainly because
these three methods are able to preserve local neighborhood structure while, to some extent, preventing data points from crowding together.
It is promising that all of these methods perform  better than an exhaustive $\ell_2$ scan using the uncompressed GIST features. 

Figure~\ref{Fig:EmbeddingSelection} shows that
 LE (\IDH-LE$_\BB$ in the figure), the most widely used embedding method in hashing, does not perform as well as a variety of other methods (including t-SNE), and in fact performs worse than PCA, which is a linear technique.
This is not surprising because LE (and similarly LLE) tends to collapse large portions of the data (and not only nearby samples in the original space) close together in the low-dimensional space.
The results are consistent with the analysis in \cite{tSNE2008,EE2010}.
Based on the above observations, we argue that  manifold learning methods (\eg t-SNE, EE), which not only preserve local similarity  but also force dissimilar data apart in the low-dimensional space, are more effective than the popular LE for hashing.

It is interesting to see that \IDH-PCA outperforms PCAH \cite{SSH2012}
by a large margin, despite the fact that PCAH is performed on the
whole training data set. This shows that the generalization capability of \IDH
based on a very small set of data points also  works for  linear
dimensionality methods.

\section{Experimental results}
We evaluate \IDH on four large scale image datasets: CIFAR-10\footnote{\url{http://www.cs.toronto.edu/~kriz/cifar.html}}, MNIST,
SIFT1M \cite{SSH2012} and GIST1M\footnote{\url{http://corpus-texmex.irisa.fr/}}.
The MNIST dataset consists of $70,000$ images, each of 784 dimensions, of handwritten digits from `0' to `9'.   As a subset of the well-known 80M tiny image collection \cite{80Mtiny2008}, CIFAR-10 consists of 60,000 images which are manually labelled as 10 classes with $6,000$ samples for each class. We represent each image in this dataset by a GIST feature vector \cite{GIST2001} of dimension $512$. For MNIST and CIFAR-10, the whole dataset is split into a test set with $1,000$ samples and a training set with all remaining samples.

We compare nine hashing algorithms including the proposed \IDH-tSNE, \IDH-LE and seven other unsupervised state-of-the-art methods: PCAH \cite{SSH2012}, SH \cite{SH08}, AGH \cite{AGH2011} and STH \cite{STH2010}, BRE \cite{BRE2009}, ITQ \cite{PCA-ITQ}, Spherical Hashing (SpH) \cite{spherical2012}.  
We use the provided codes and suggested parameters according to the authors of these methods. 
Because our methods are fully unsupervised we did not consider supervised methods in our experiments.
Due to the high computational cost of BRE and high memory cost of STH, we sample $1,000$ and $5,000$ training points for these two methods respectively.
We measure  performance by mean of average precision (MAP) or
precision and recall curves for \textit{hamming ranking} using 16 to
128 hash bits. We also report the results  for \textit{hash lookup}
using a Hamming radius within 2 by F1 score \cite{F108}: $F_1 = 2
( precision \cdot recall) / (precision + recall)$. Ground truths are
defined by the category information for the labeled datasets MNIST
and CIFAR-10, and by Euclidean neighbors for SIFT1M and GIST1M.

\begin{figure}
 \centering
 \includegraphics[width = 0.37\textwidth]{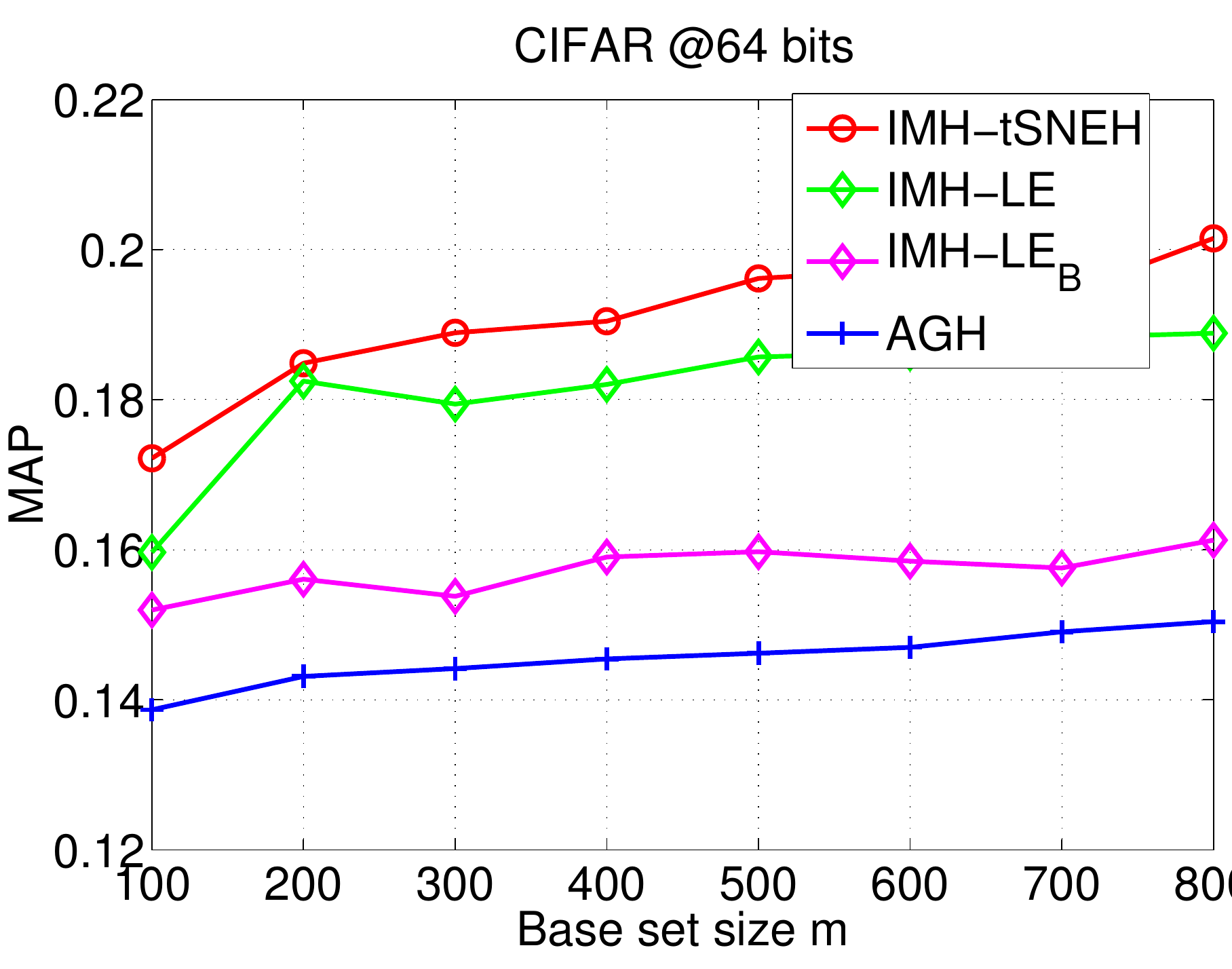}
 \\
 \includegraphics[width = 0.37\textwidth]{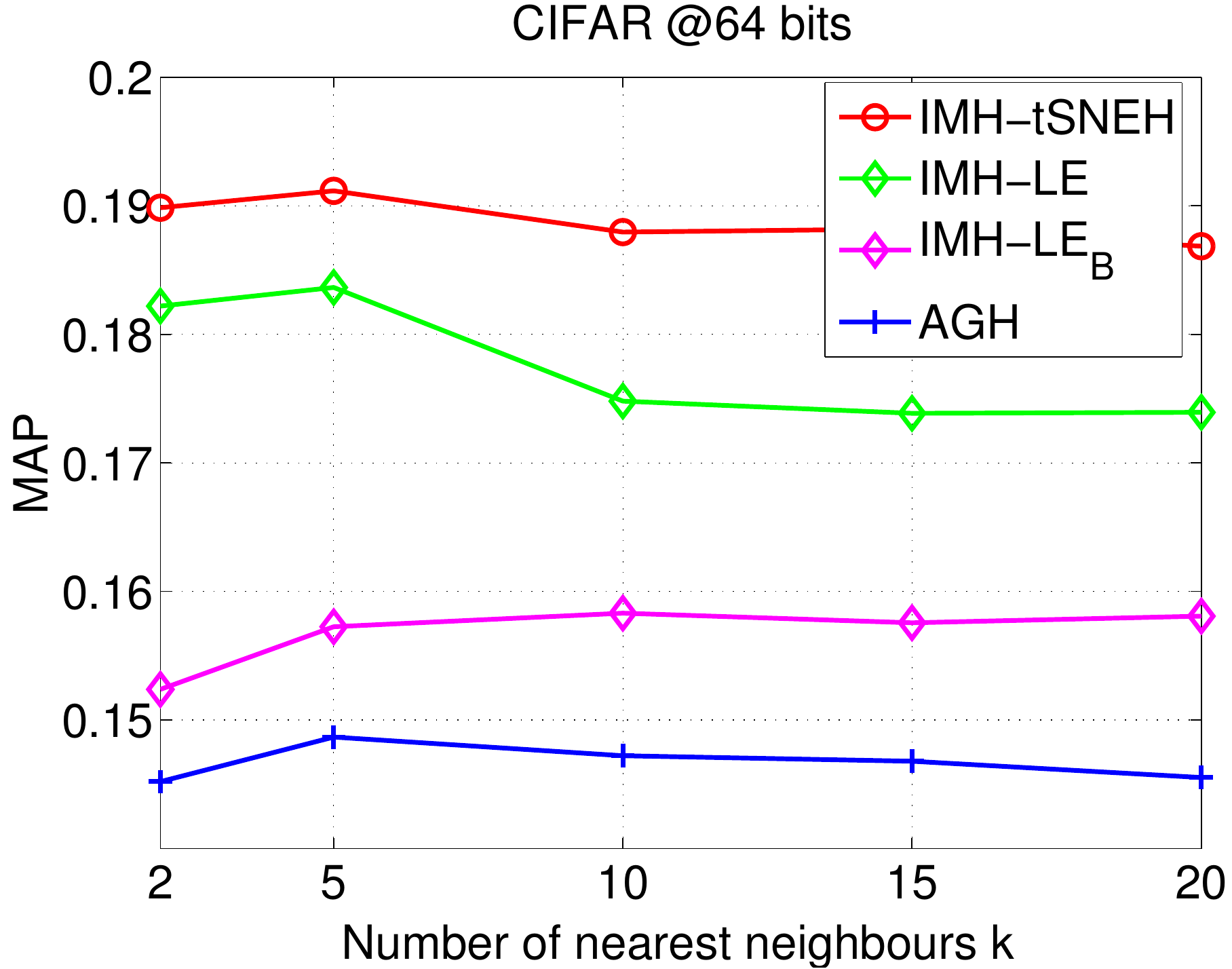}
\label{compare_anchor_num}
\vspace{-0.3cm}
\caption{MAP results versus varying  base set size $m$ (left, fixing $k = 5$) and  number of nearest base points $k$ (right, fixing $m = 400$) for the proposed methods and AGH.
The comparison is conducted on the CIFAR-10 dataset using 64-bits .}
\label{Fig:compare_anchornum}
\end{figure}
\begin{table}[]
\centering
 \resizebox{0.4\textwidth}{!}{
\begin{tabular}{cc|cccc}
\hline\hline
bits & & \IDH-LE$_\BB$ & \IDH-LE & \IDH-tSNE & AGH\\
 \hline
\multirow{2}{*}{32} &
 Random & 14.07 & 16.20 & 17.26 & -\\
 &K-means & \textbf{16.05} & \textbf{17.48} & \textbf{18.38} & 15.76\\
 \hline
 \multirow{2}{*}{64} &
 Random & 14.64 & 16.98 & 16.93 & - \\
 &K-means & \textbf{15.90} & \textbf{18.20} & \textbf{19.0}4 & 14.55\\
 \hline
 \multirow{2}{*}{96} &
 Random & 14.76 & 17.02 & 17.21 & -\\
 &K-means & \textbf{15.46} &  \textbf{18.56} & \textbf{19.41} & 13.98\\
 \hline\hline
\end{tabular}
}
\caption{MAP (\%) evaluation  of different base generating methods: random sampling vs. K-means.
The comparison is performed on the CIFAR-10 dataset  with code lengths from 32 to 96 and base set size 400. }
\label{compare_random_kmeans}
\end{table}

\paragraph{Base selection}
In this section,  we take the CIFAR-10 dataset for example to compare different base generation methods and different base sizes for the proposed methods. AGH is also evaluated here for comparison.
Table~\ref{compare_random_kmeans} compares two methods for generating base point sets: random sampling and K-means on the training data. Not surprisingly, we see that the performance of our methods using K-means is better at all code lengths than that using random sampling.
Also we can see that, even with base set by random sampling, the proposed methods outperform AGH in all cases but one. Due to the superior results and high efficiency in practice, we generate the base set by K-means in the following experiments.

From Figure~\ref{Fig:compare_anchornum},
we  see that the performance of the proposed methods and AGH  do not change significantly with both the base set size $m$ and the number of nearest base points $k$. Based on this observation, for the remainder of this paper, we set $m = 400$ and $k = 5$ for our methods, unless otherwise specified.
Also it is clear that \IDH-LE$_\BB$, which only enforces smoothness in the base set, does not perform as well as \IDH-LE, which also enforces smoothness between the base set and training set. Note, however, that \IDH-LE$_\BB$ is still better than AGH on this dataset.

\begin{figure}[t]
\centering
\includegraphics[width = 0.37\textwidth]{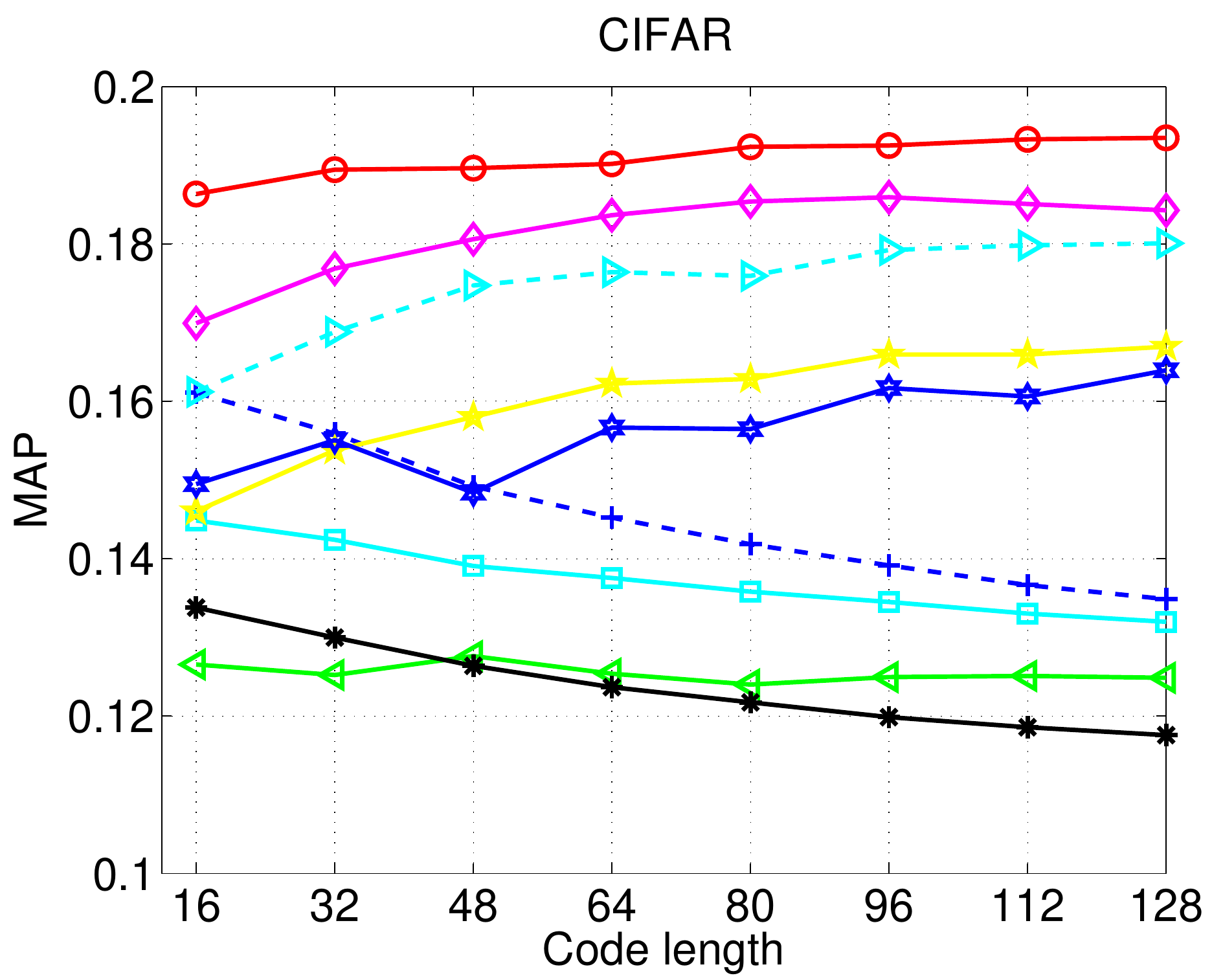}
\includegraphics[width = 0.37\textwidth]{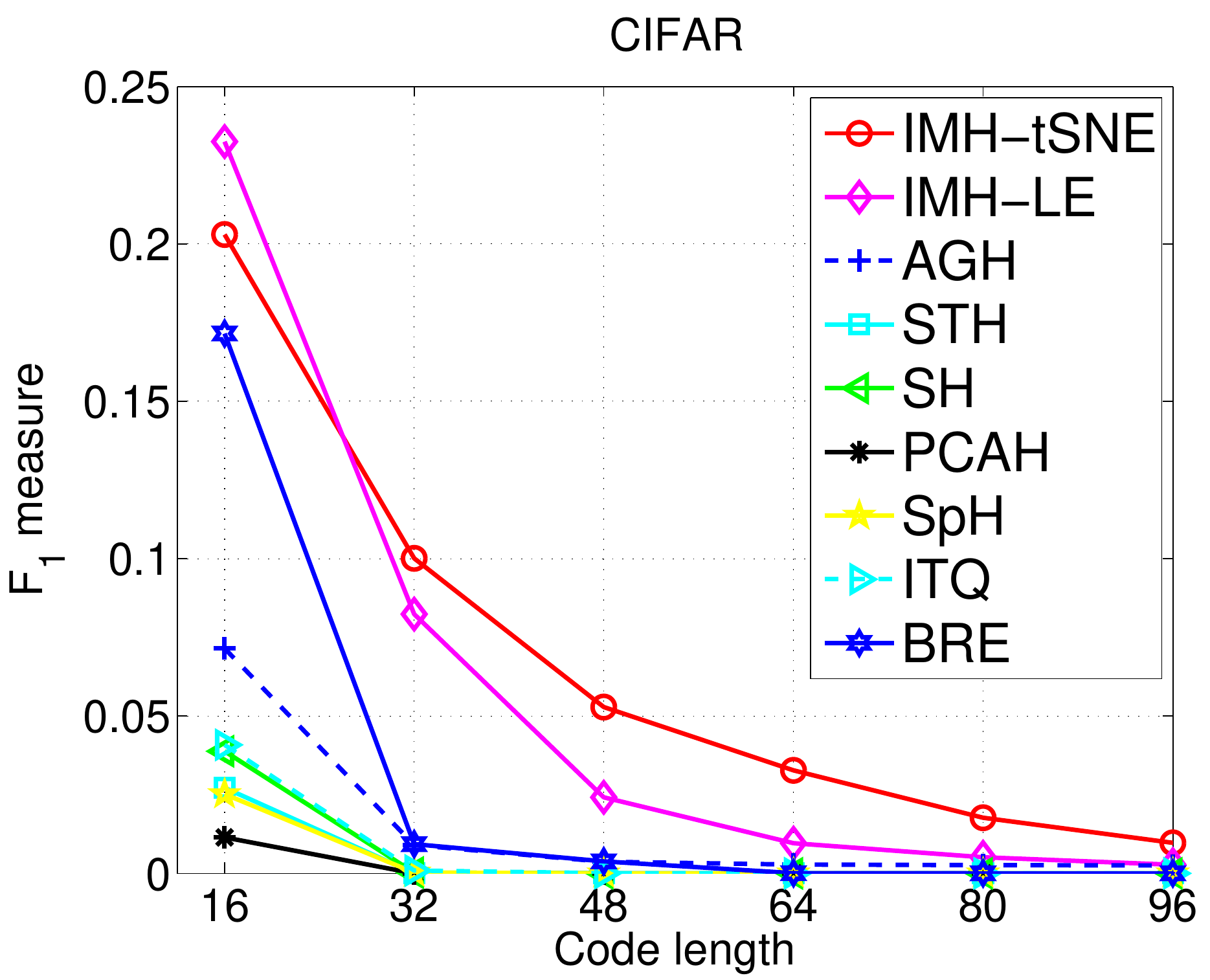}
\caption{Comparison of different methods on CIFAR-10 based on MAP (left) and $F_1$  (right) for varying code lengths.
}
\label{cifar}
\end{figure}

\paragraph{Results on CIFAR-10 dataset}
We report the comparative results based on  MAP   for  hamming ranking  with code lengths from 16 to 128 
bits 
in Figure~\ref{cifar}.
We see that the proposed \IDH-LE and \IDH-tSNE perform best in all cases. 
Among the proposed algorithms, the LE based \IDH-LE is inferior to the t-SNE based \IDH-tSNE. 
 \IDH-LE is still much better than AGH and STH, however. ITQ performs better than SpH and BRE on this dataset, but is still inferior to \IDH. 
  SH and PCAH perform worst in this case, because SH 
relies upon its uniform data 
assumption while PCAH simply generates the hash hyperplanes by PCA directions, which does not explicitly capture the similarity information. The results are consistent with the complete precision and recall curves shown in the supplementary material.
We also report the $F_1$ results for hash lookup with Hamming radius 2
It is can be seen that \IDH-LE and \IDH-tSNE also outperform all other methods by large margins.  BRE and AGH obtain  better results than the remaining methods, although 
the performance of all methods drop as code length grows.

\begin{figure}
    \centering
\includegraphics[width = 0.37\textwidth]{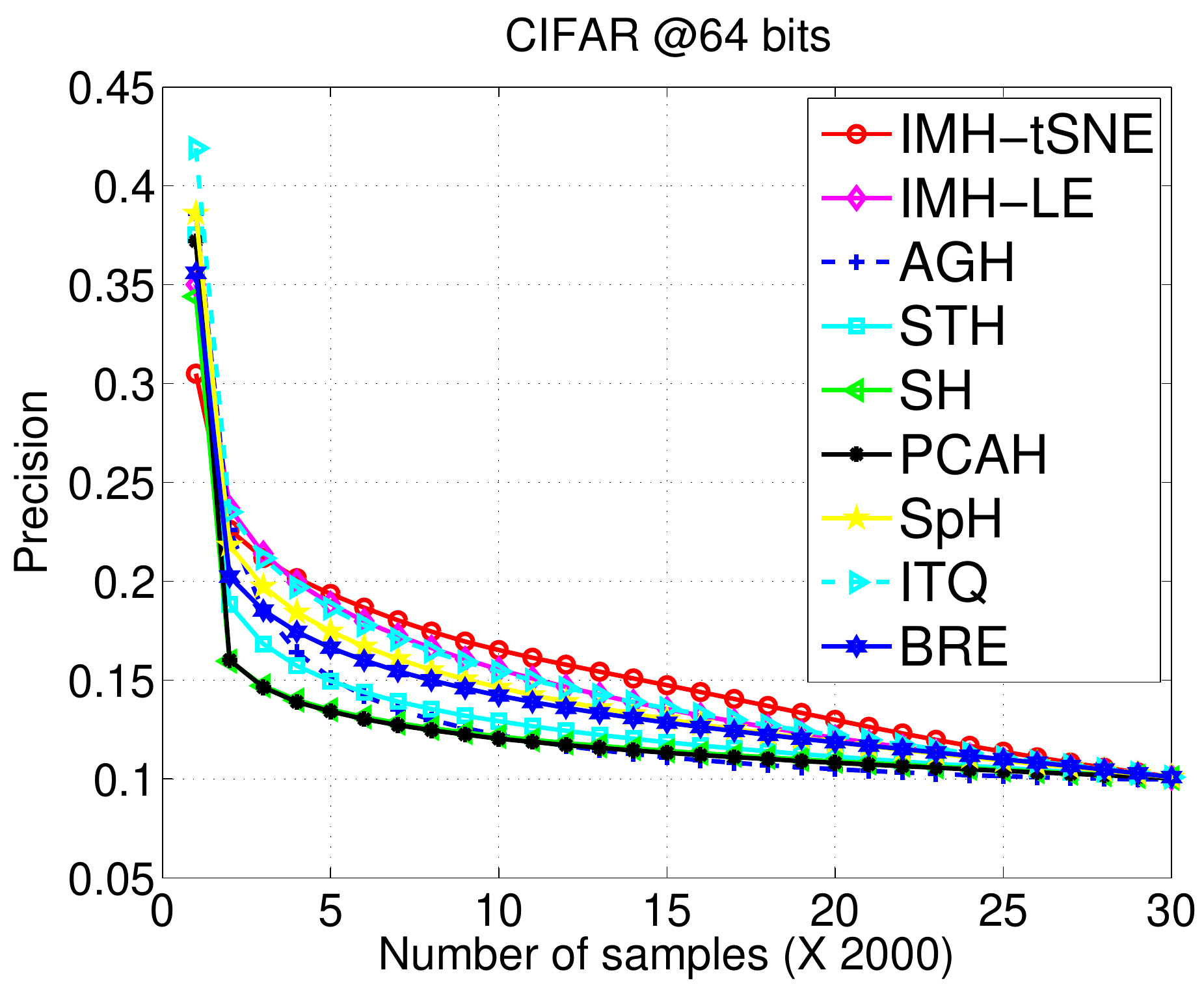}
\includegraphics[width = 0.37\textwidth]{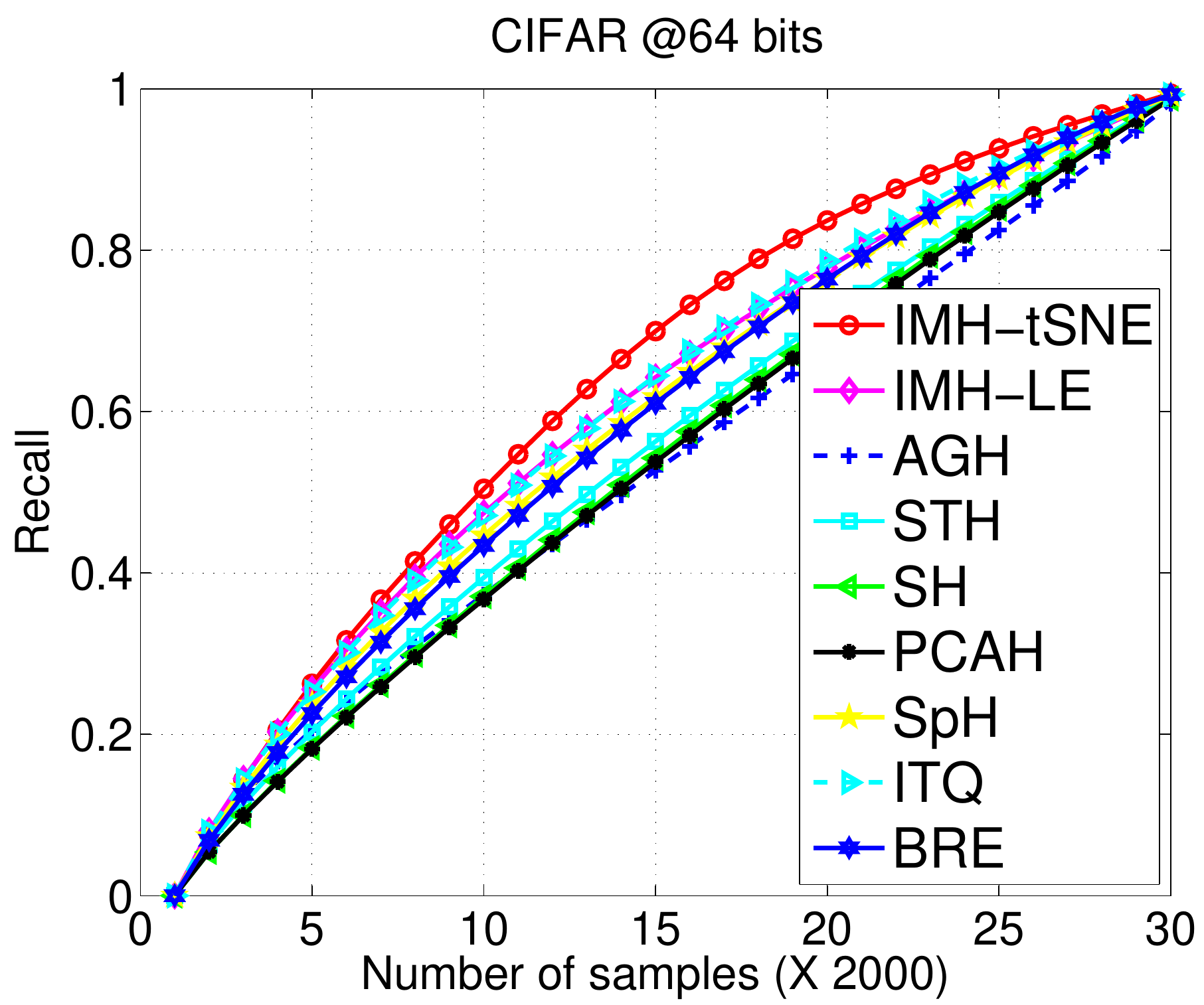}
\caption{Comparison of different methods on CIFAR-10 based on precision (left) and recall (right) using 64-bits.
Please refer to the complementary for complete results for other code lengths.}
\label{cifar_pr}
\end{figure}

Figure~\ref{cifar_pr} shows the  precision and recall curves of hamming ranking for the compared methods. We see that STH and AGH obtain relatively high precisions when a small number of samples are returned, however precision drops significantly as the number of retrieved samples increases. In contrast,
 \IDH-tSNE, \IDH-LE and ITQ achieve  higher precisions with relatively larger numbers of retrieved points. 

\begin{figure*}[t]
\centering
\begin{subfigure}[a]{0.06\textwidth}
\includegraphics[width = \textwidth]{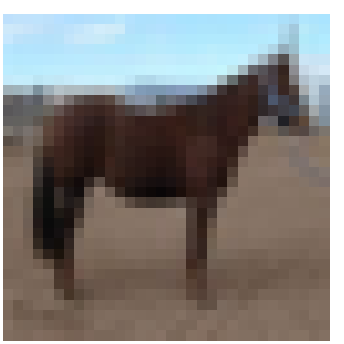}
\caption{Query}
\end{subfigure}
\quad
\begin{subfigure}[c]{0.15\textwidth}
\includegraphics[width = \textwidth]{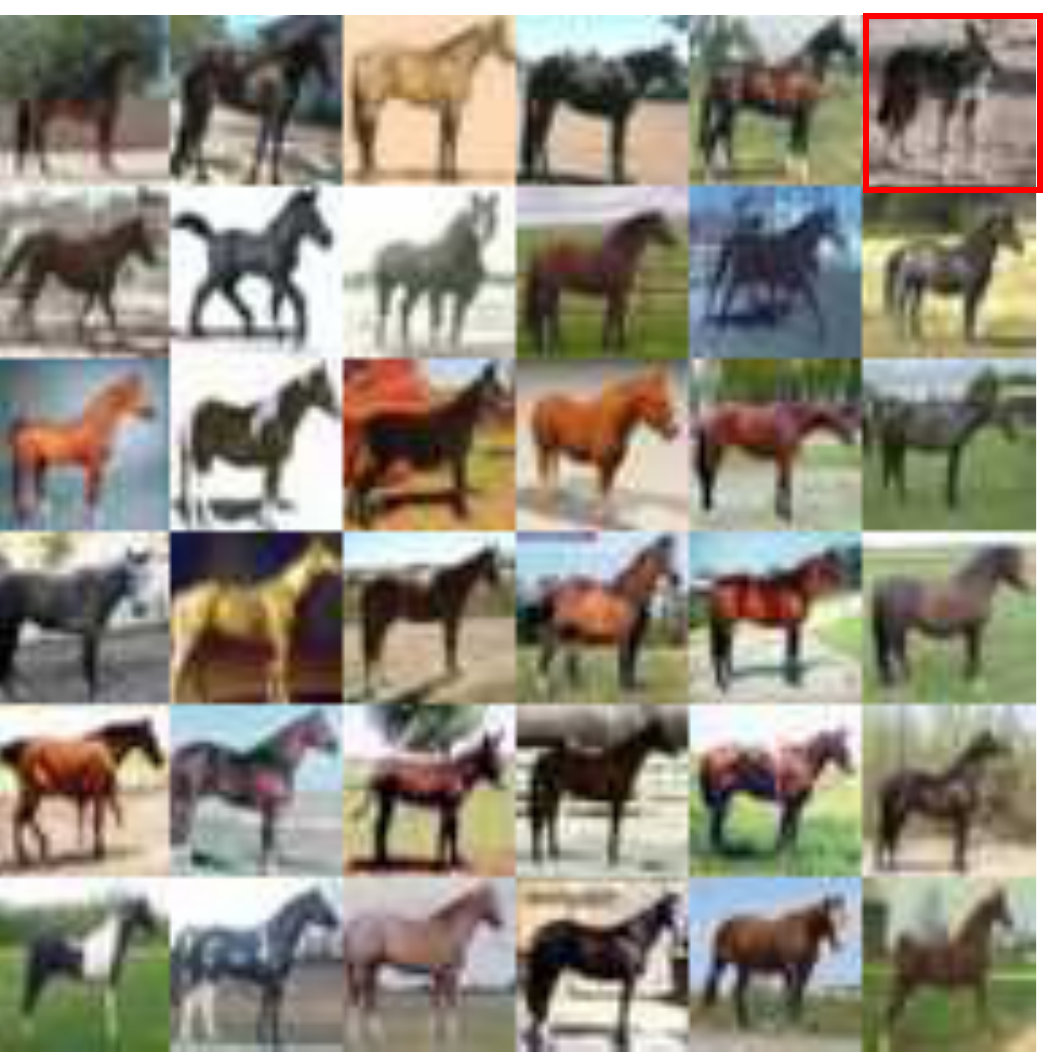}
\caption{\IDH-tSNE}
\end{subfigure}%
\quad
\begin{subfigure}[c]{0.15\textwidth}
\includegraphics[width = \textwidth]{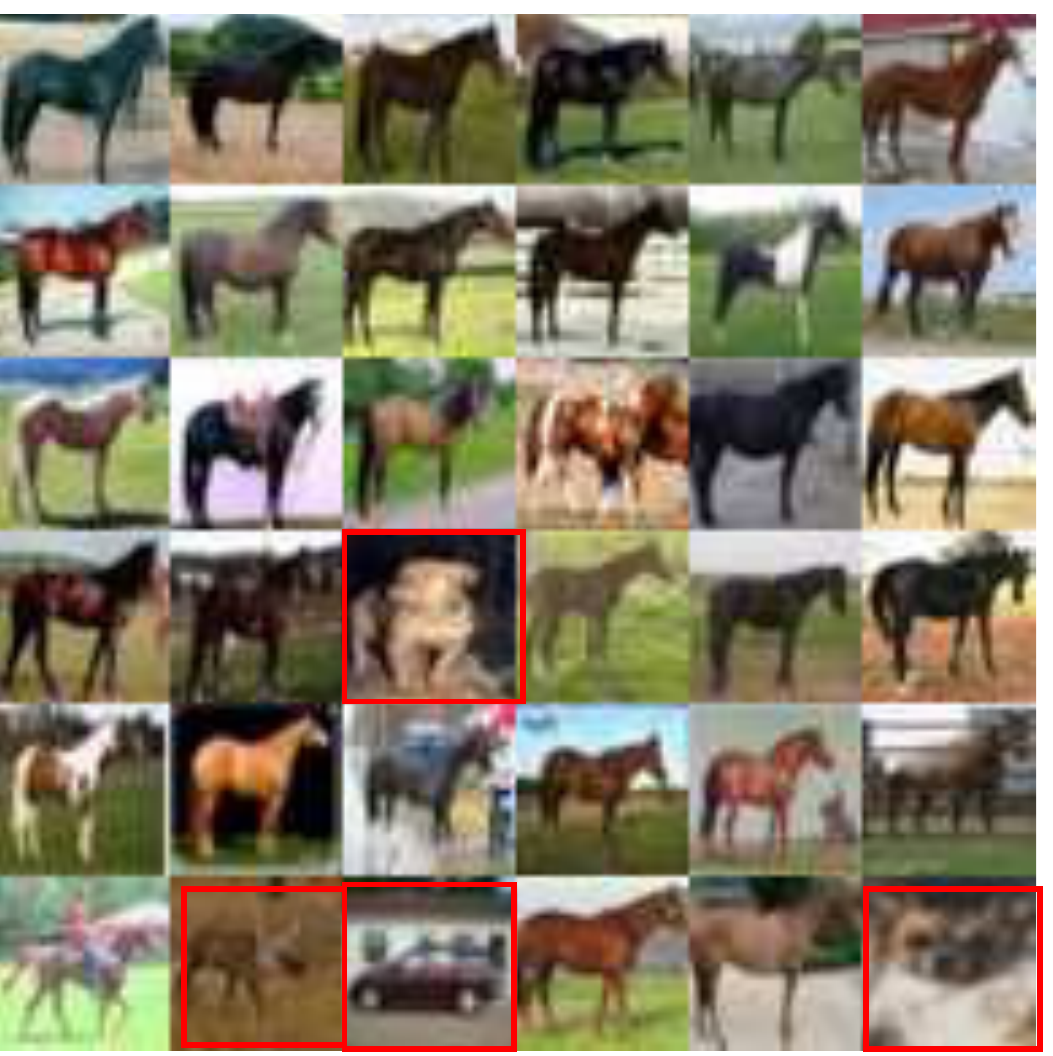}
\caption{SH}
\end{subfigure}%
\quad
\begin{subfigure}[d]{0.15\textwidth}
\includegraphics[width = \textwidth]{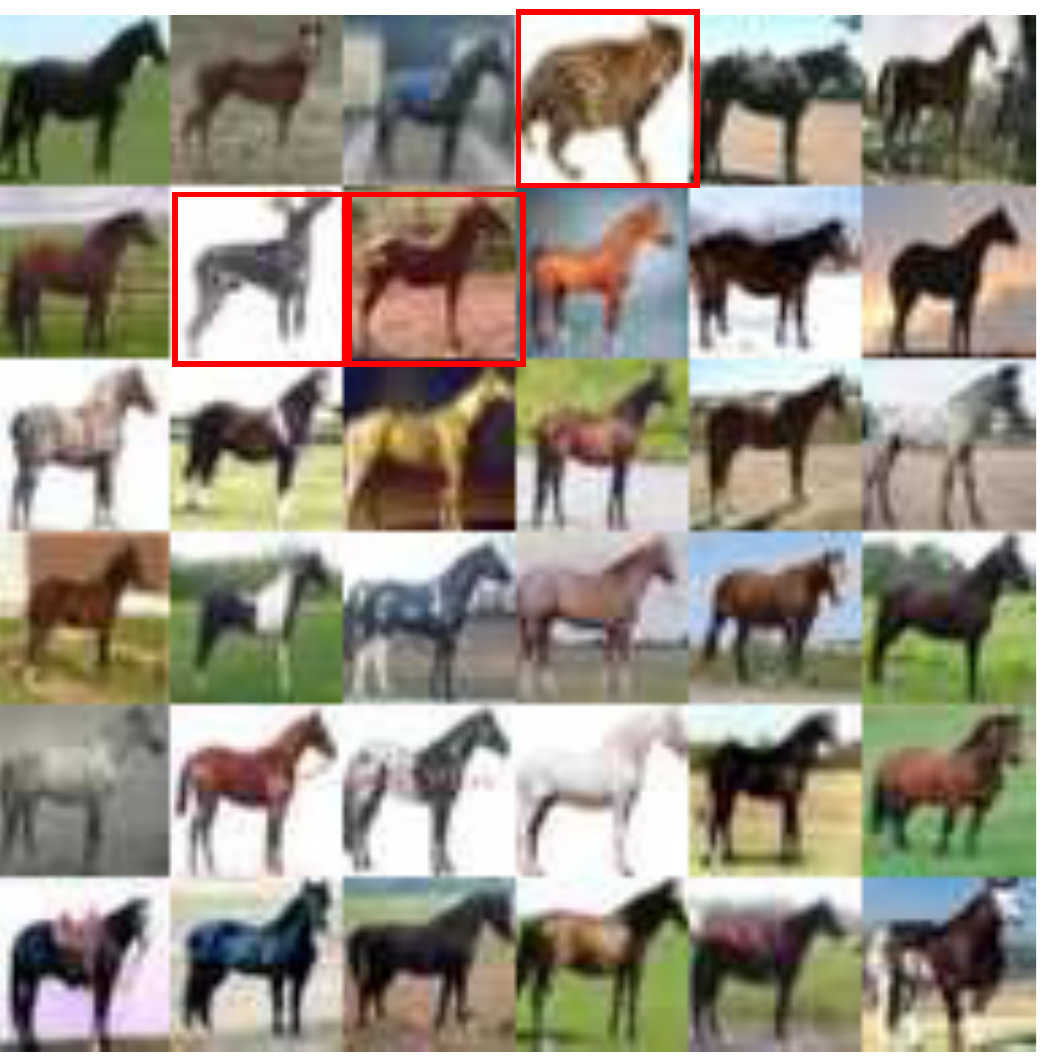}
\caption{AGH}
\end{subfigure}%
\quad
\begin{subfigure}[e]{0.15\textwidth}
\includegraphics[width = \textwidth]{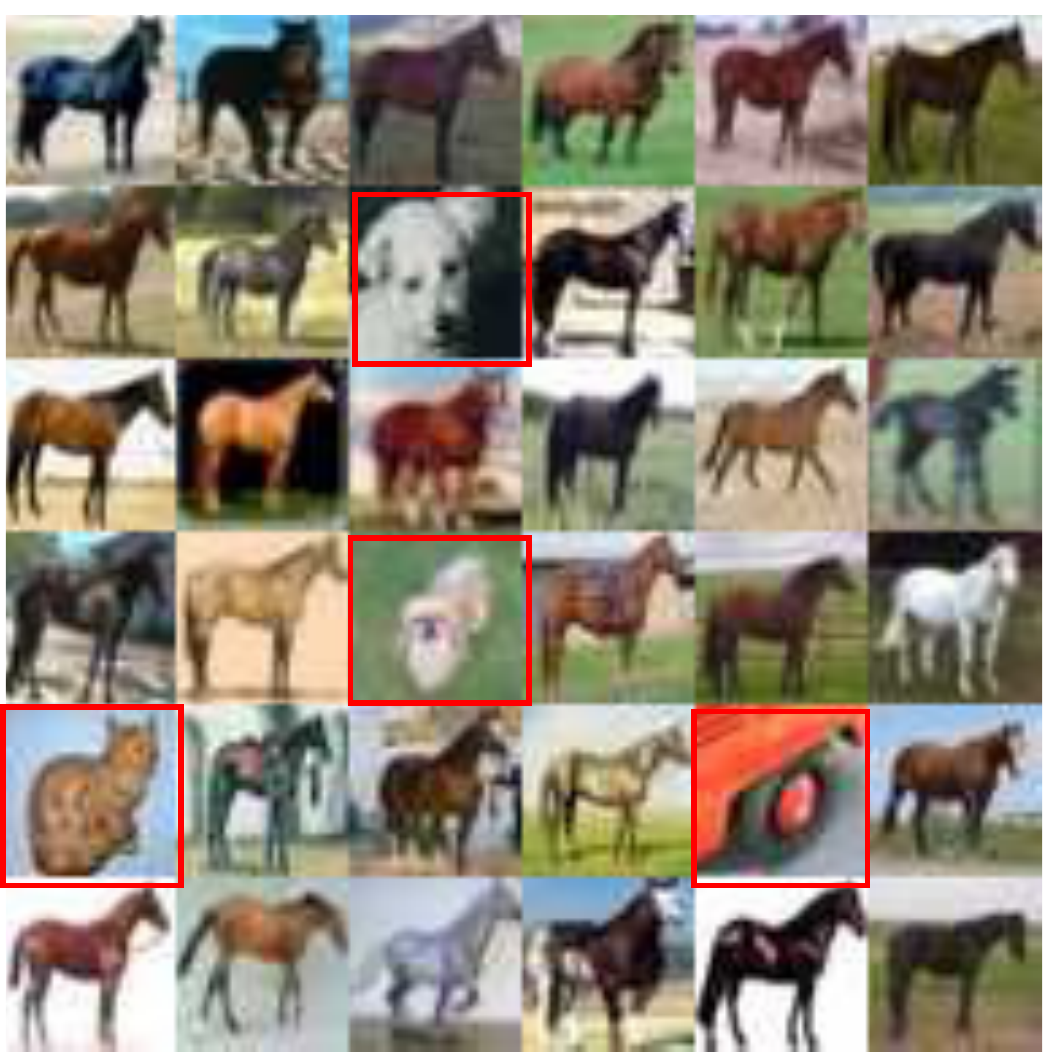}
\caption{STH}
\end{subfigure}%
\vspace{10px}
\caption{The query image (a) and the query results returned by various methods with 32 hash bits. False positive returns are marked with red borders.}
\label{Fig:returned_images}
\end{figure*}

We also show qualitative results of \IDH and related methods on a sample query in Figure~\ref{Fig:returned_images}.
As can be seen, \IDH-tSNEH achieves the best search quality in term of visual relevance.

\begin{figure}[b!]
\centering
\includegraphics[width = 0.35\textwidth]{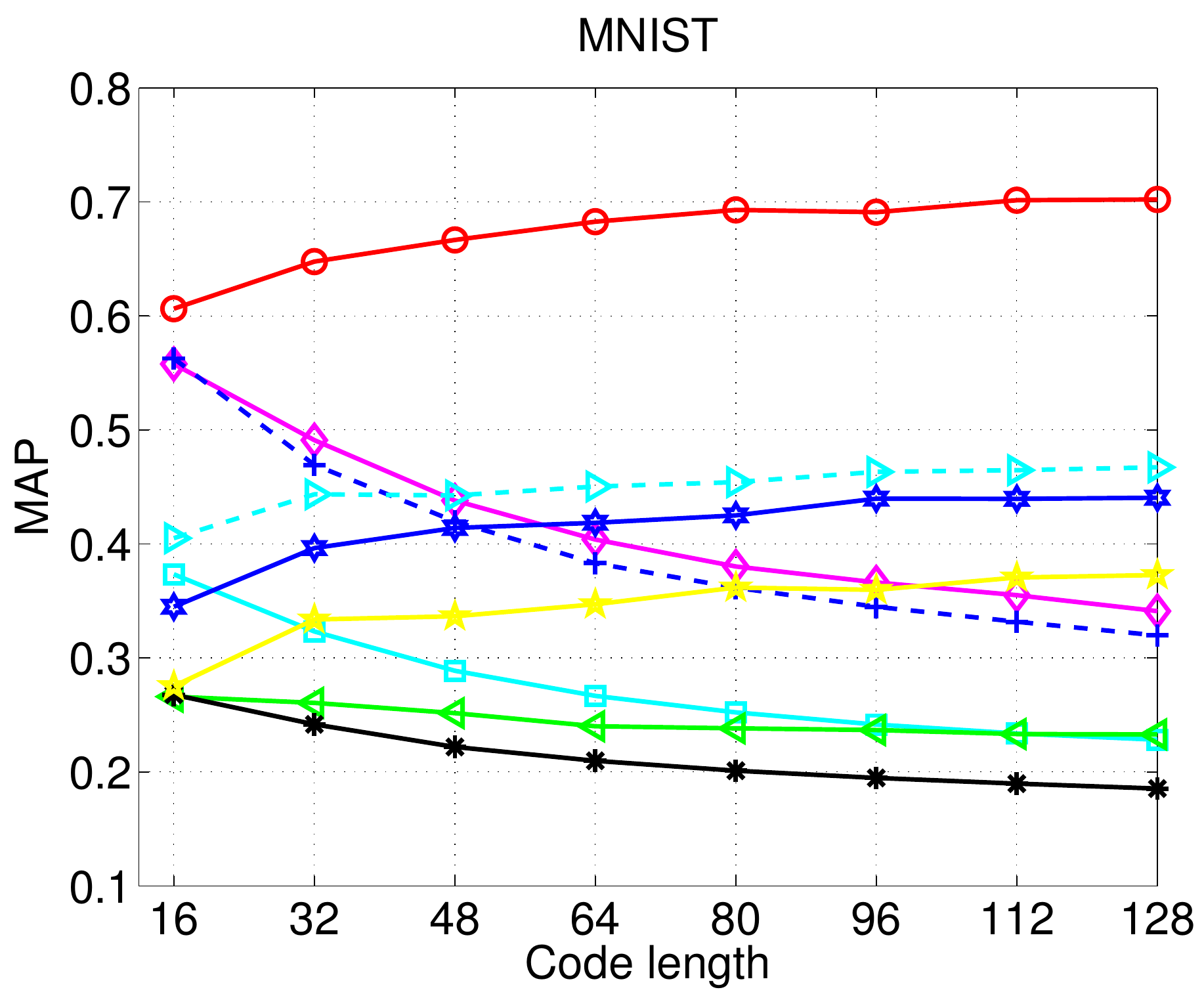}
\\
\includegraphics[width = 0.35\textwidth]{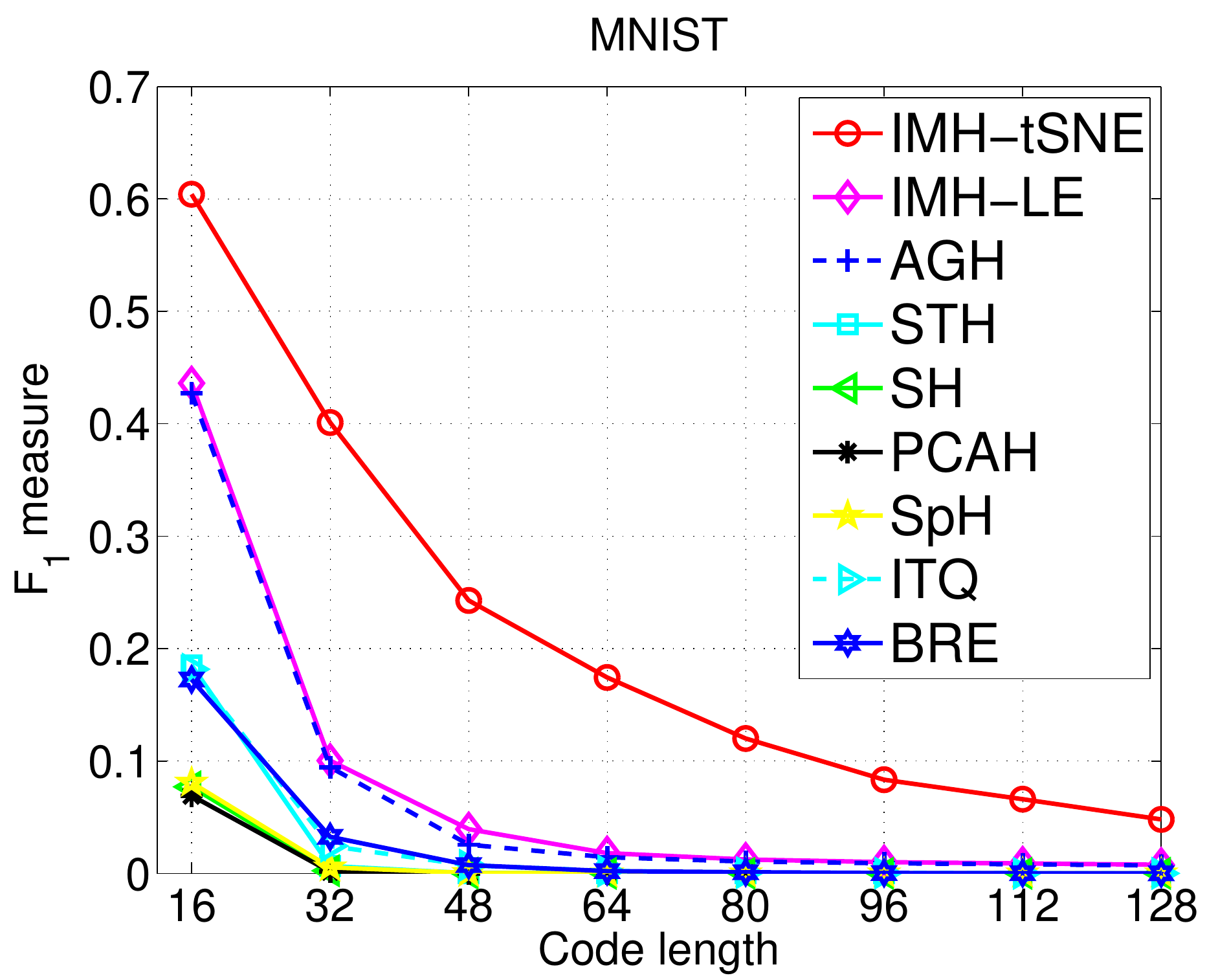}
\caption{Comparison of different methods on the MNIST dataset using MAP (left) and $F_1$  (right) for varying code lengths.
}
\label{mnist}
\end{figure}

\begin{table}[b]
\centering
\resizebox{0.38\textwidth}{!}{
\begin{tabular}{ccc|cc}
\hline\hline
Method & \multicolumn{2}{c|}{Train time} & \multicolumn{2}{c}{ Test time}\\
&64-bits  & 128-bits  & 64-bits  & 128-bits \\
\hline
\IDH-LE&9.9 & 9.9& $5.1 \times 10^{-5}$& $3.8 \times 10^{-5}$ \\
\IDH-tSNE&16.7&20.2 &$2.8 \times 10^{-5}$ & $3.1 \times 10^{-5}$\\
SH&6.8 & 16.2&$5.8 \times 10^{-5}$ &$1.8 \times 10^{-4}$  \\
STH&266.1& 485.4& $1.8\times 10^{-3}$& $3.6 \times 10^{-3}$ \\
AGH&9.5 & 9.5&$4.7 \times 10^{-5}$ &$5.5 \times 10^{-5}$ \\
PCAH&3.8 &4.1 &$5.7 \times 10^{-6}$ &$1.2 \times 10^{-5}$\\
SpH & 19.7&41.0 &$1.3 \times 10^{-5}$ & $2.0 \times 10^{-5}$\\
ITQ &10.4 &20.3 &$6.9 \times 10^{-6}$ & $1.1 \times 10^{-5}$\\
BRE &418.9 &1731.9 &$1.2\times 10^{-5}$ &$2.4 \times 10^{-5}$ \\
\hline\hline
\end{tabular}
}
\caption{Comparison of training and testing times (in seconds) on MNIST with 70K 784D feature points. K-means dominates the cost of  AGH and \IDH (8.9 seconds), which can be conducted in advance in practice. The experiments are based on a desktop PC with a 4-core 3.07GHZ CPU and 8G RAM.}
\label{Tab:time_mnist}
\end{table}
\paragraph{Results on MNIST dataset}
The MAP and $F_1$ scores for these compared methods are reported in Figure~\ref{mnist}. As in Figure~\ref{cifar}, \IDH-tSNE achieves the best results. On this dataset we can clearly see that \IDH-tSNE outperforms \IDH-LE by a large margin, which increases as code length increases. 
This further demonstrates the advantage of t-SNE as a tool for hashing by embedding high dimensional data into a low dimensional space. The dimensionality reduction procedure  not only preserves the local neighborhood structure, but also reveals important global structure (such as clusters) \cite{tSNE2008}. Among the four LE-based methods, while \IDH-LE shows a small advantage over AGH, both methods achieve much better results than STH and SH. ITQ and BRE obtain high MAPs with longer bit
lengths,
but they still perform less well for the hash look up $F_1$.
 PCAH  performs worst  in terms of both MAP and the $F_1$ measure.  Refer to the supplementary material for the complete precision and recall curves which validate the observations here.

\textit{Efficiency} %
Table~\ref{Tab:time_mnist} shows training and testing time on the MNIST dataset for various methods, and shows that
the linear method, PCAH, is fastest.
 \IDH-tSNE is slower than \IDH-LE, AGH and SH in terms of training time, however all of these methods have relatively low execution times and are much faster than STH and BRE. In terms of test time, both \IDH algorithms are comparable to other methods,  except STH which takes much more time to predict the binary codes by SVM on this non-sparse dataset.

\begin{figure}[t!]
\centering
\includegraphics[width = 0.35\textwidth]
{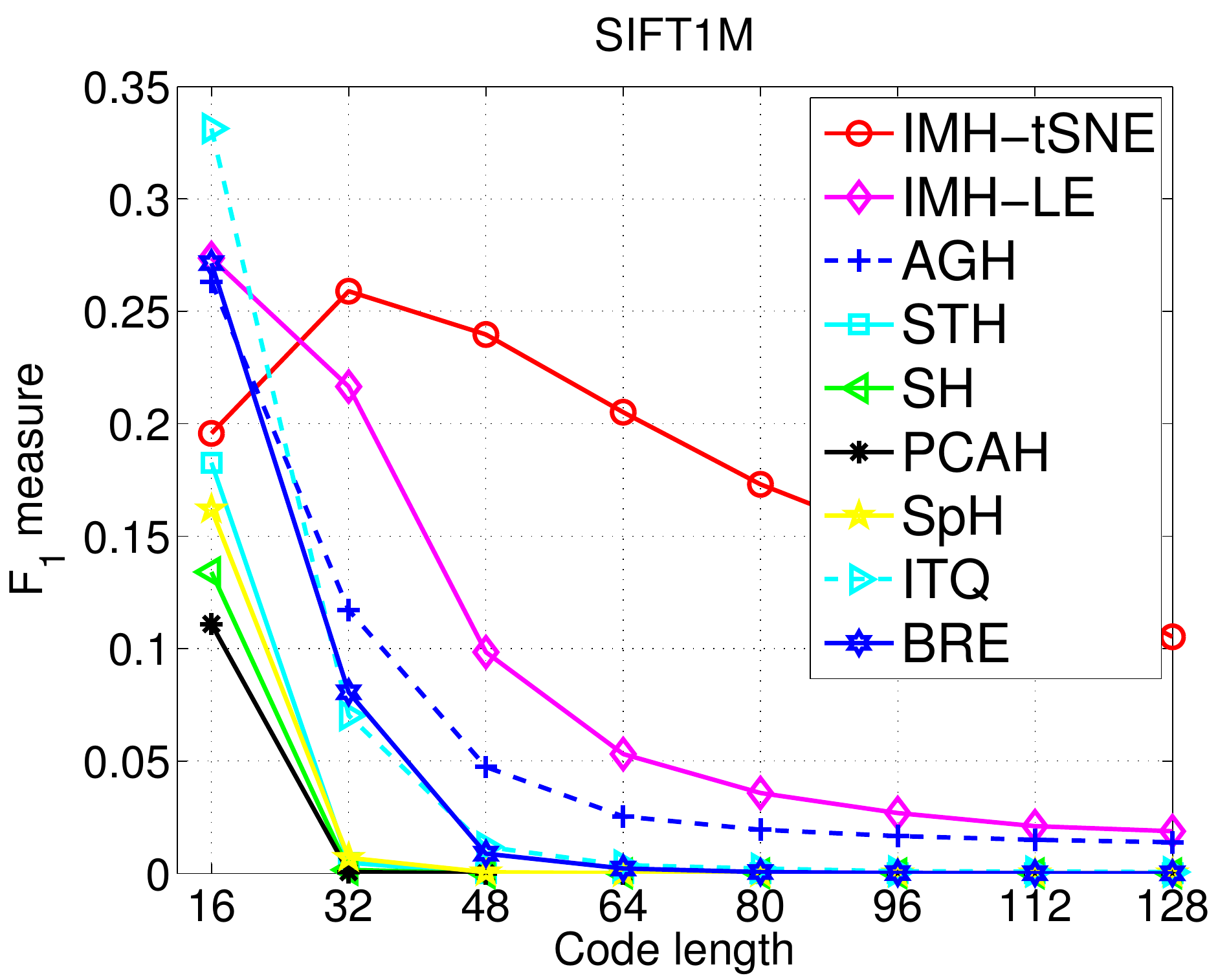}
\includegraphics[width = 0.35\textwidth]
{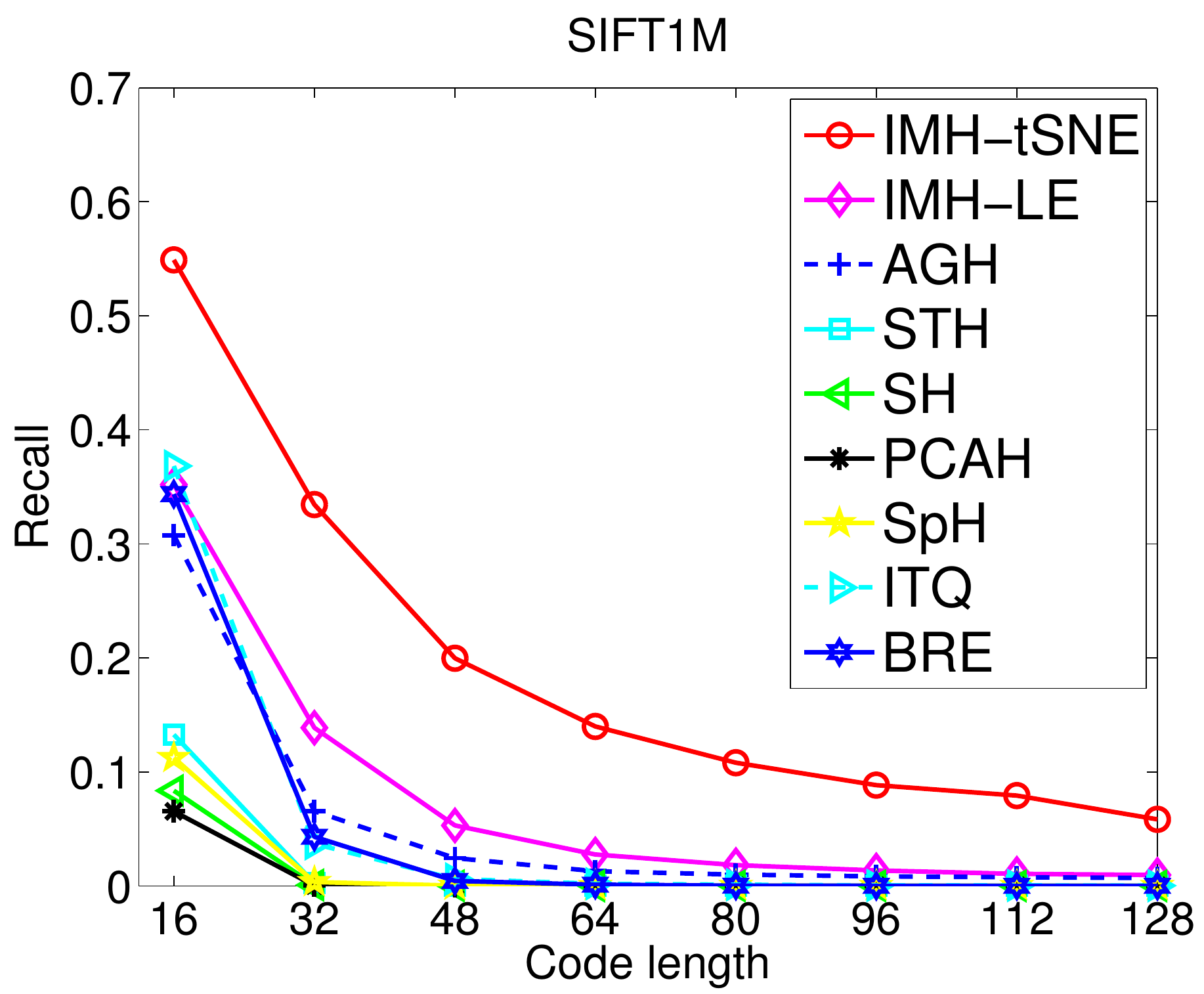}
\caption{Comparative results results on SIFT1M for $F_1$ (left) and recall (right) with hamming radius 2. Ground truth is defined to be the closest 2 percent of points as measured by the Euclidean distance.}
\label{sift1M}
\end{figure}

\begin{figure}[b!]
\centering
\includegraphics[width = 0.35\textwidth]
{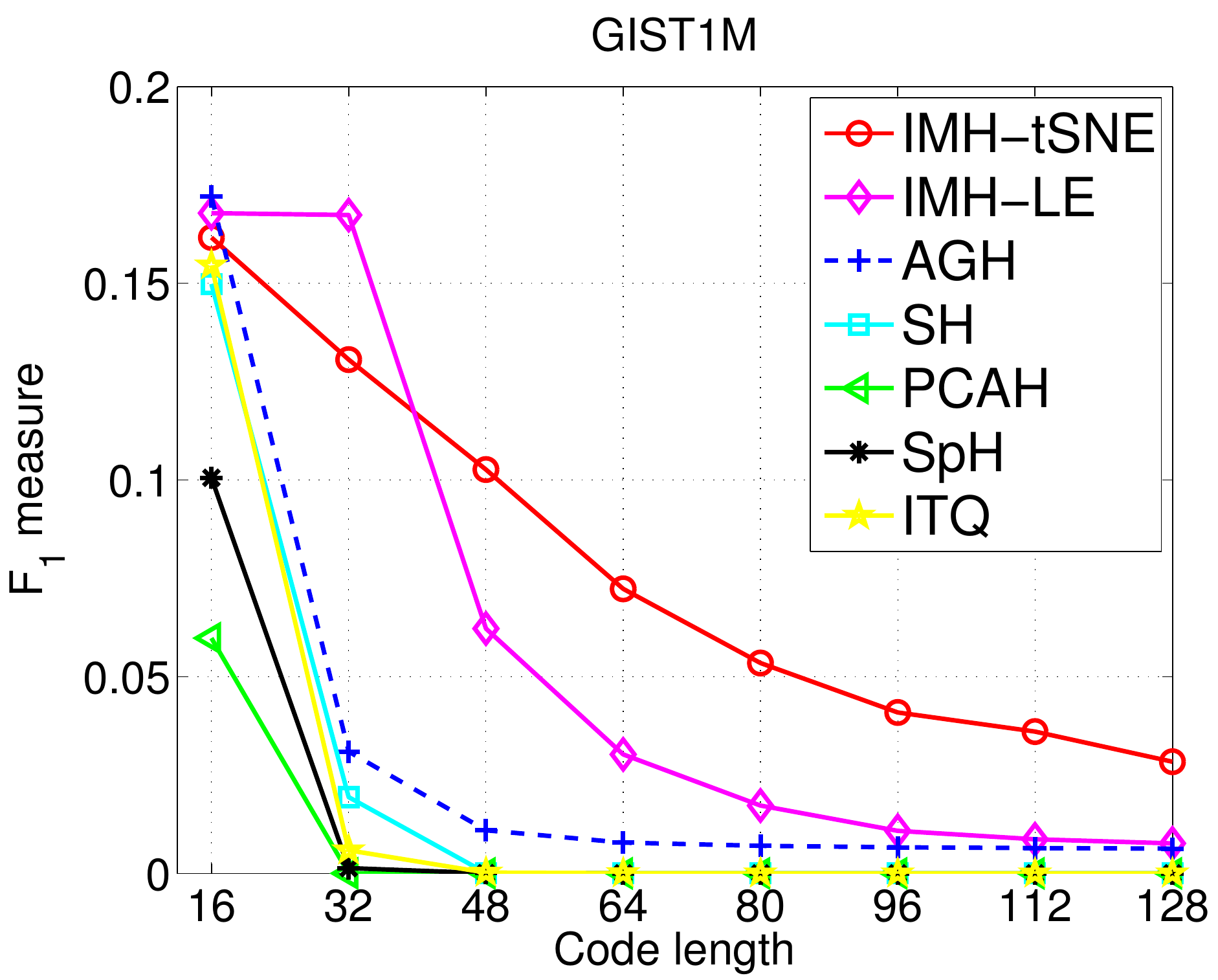}
\includegraphics[width = 0.35\textwidth]
{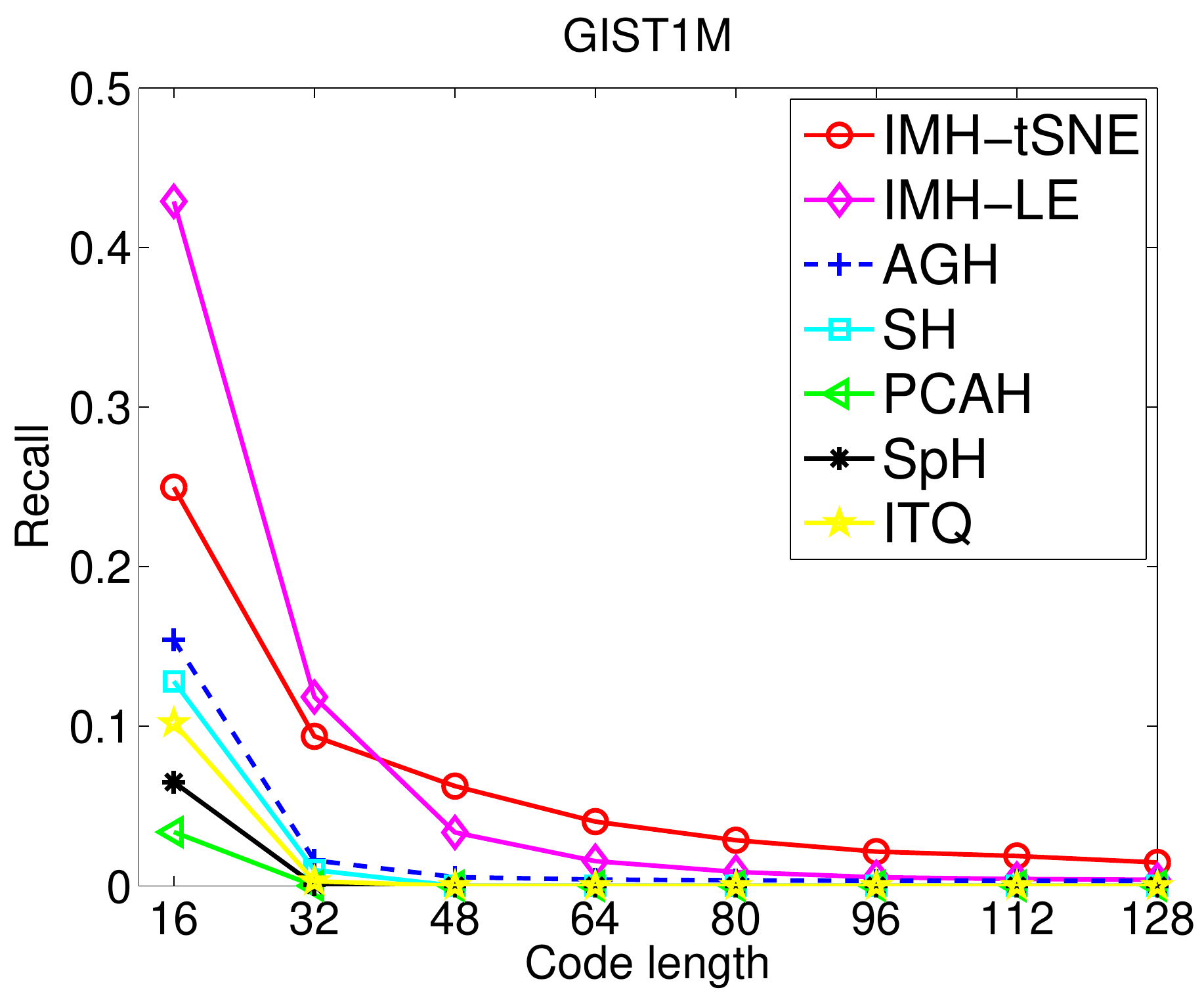}
\caption{Comparative results results on GIST1M by  $F_1$ (left) and recall (right) with hamming radius 2. Ground truth is defined to be the closest 2 percent of points as measured by the Euclidean distance.}
\label{GIST1M}
\end{figure}

\paragraph{Results on SIFT1M and GIST1M}
SIFT1M  contains one million local SIFT descriptors extracted from a large set of images \cite{SSH2012}, each of which is represented by a 128D vector of histograms of gradient orientations. 
GIST1M contains one million GIST features and
each feature is represented by a 960D vector. 
 For both of these datasets, one million samples are used as training set and additional 10K are used for testing. As in \cite{SSH2012}, ground truth is defined as the  
closest 2 percent of points as measured by the
Euclidean distance. For these two large datasets, we generate $1,000$ points by K-means and set $k = 2$ for both \IDH and AGH.
The comparative results on SIFT1M and GIST1M are summarized in Figure~\ref{sift1M} and Figure~\ref{GIST1M}, respectively.
Again, \IDH consistently achieves superior 
results in terms of both $F_1$ score and recall with hamming radius 2. We see that the performance of most of these methods decreases dramatically with increasing code length as the hamming spaces become more sparse, which makes the hash lookup fail more often. However \IDH-tSNE still achieves relatively high scores with large code lengths. If we look at Figure~\ref{sift1M} (left), ITQ obtains the highest $F_1$ with 16-bits, however it decreases to near zero at 64-bits. In contrast, \IDH-tSNE still manages an $F_1$ of $0.2$.  Similar results are observed in the recall curves.

\begin{figure}[t!]
\centering
\includegraphics[width = 0.5\textwidth]{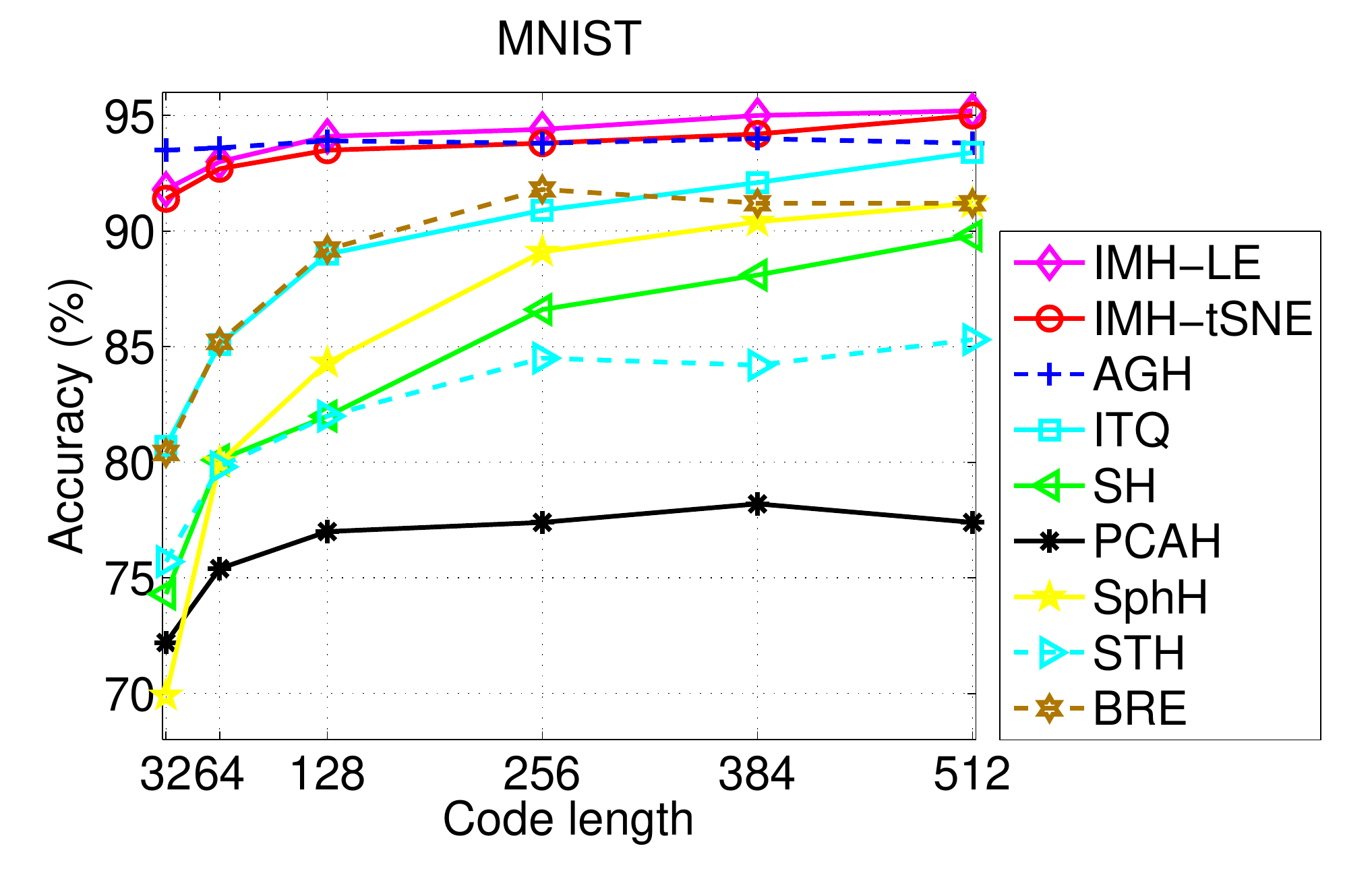}
\caption{Classification accuracy (\%) on MNIST with binary codes of various hashing methods by linear SVM.}
\label{calssification_mnist}
\end{figure}

\paragraph{Classification on binary codes}
In order to demonstrate classification performance 
we have trained a linear SVM on the binary codes generate by \IDH for the MNIST data set.
In order to learn codes with higher bit
lengths
  for \IDH and AGH, we set the size of the base set to $1,000$.
Accuracies of different binary 
encodings
are shown in Figure~\ref{calssification_mnist}. Both \IDH and AGH achieve high accuracies on this dataset, although \IDH performs better with higher code 
lengths. 
In contrast, the best results of all other methods, obtained by ITQ,
are consistently worse than those for \IDH, especially for short code
lengths. Note that even with only 128-bit binary features \IDH obtains
a high 94.1\%. Interestingly, we get the same classification rate of
94.1\% applying the linear SVM to the uncompressed 784D features, which occupy several hundreds times as much space as the learned hash codes.

\section{Conclusion}

We have proposed a simple yet effective hashing framework which
provides a practical connection between manifold learning methods
(typically non-parametric and with high computational cost) and hash
function learning (requiring high efficiency).  By preserving the
underlying manifold structure with several non-parametric
dimensionality reduction methods, the proposed
hashing
 methods outperform several state-of-the-art methods 
in terms of 
both hash lookup and hamming ranking on several large-scale
retrieval-datasets.  The proposed inductive formulation of the hash
function sees the proposed methods require only linear time
($\mathit{O}(n)$) for indexing all of the training data and a constant
search time for a novel query.  The learned hash codes were also shown
to have  promising results on a classification problem even with  very
short code lengths.

 {\it
 This work was in part supported by ARC Future Fellowship FT120100969. 
 }

{\small
\bibliographystyle{ieee}
\bibliography{CSRef}
}

\end{document}